\documentclass{article} %
\usepackage{iclr2027_conference,times}

\usepackage{hyperref}
\usepackage{url}

\usepackage{graphicx}
\usepackage{booktabs}
\usepackage{algorithm}
\usepackage{changes}
\renewcommand{\added}[2][]{#2}
\renewcommand{\deleted}[2][]{}
\renewcommand{\replaced}[3][]{#2}
\usepackage{algpseudocode}
\usepackage[table]{xcolor}
\usepackage{array, colortbl}
\usepackage{xurl}
\usepackage{bm}
\usepackage{amssymb}
\usepackage{amsmath}
\usepackage{booktabs, array, multirow}
\usepackage{enumitem}
\usepackage{wrapfig}
\usepackage{caption}
   \hypersetup{colorlinks=true, linkcolor=blue!60!black, citecolor=blue!60!black, urlcolor=blue!60!black}
\usepackage{placeins}
\setaddedmarkup{\textcolor{blue}{#1}}
\setdeletedmarkup{\textcolor{red}{\sout{#1}}}

\newcommand{\Derek}[1]{}
\newcommand{\Saurabh}[1]{}
\newcommand{\TODO}[1]{}

\newcommand{\pihalf}{\pi_{0.5}}

\newcommand{\samethanks}[1][\value{footnote}]{\footnotemark[#1]}

\title{Fine-Tuning VLAs with Self-Demonstrated Generative Control for Multi-Task Manipulation}

\author{Prachi Garg, Steve Xing\thanks{Equal contribution.}, Prahit Yaugand\samethanks[1], Saurabh Gupta, Derek Hoiem \\
University of Illinois Urbana-Champaign \\
\texttt{\{prachig3, stevex2, yaugand2, saurabhg, dhoiem\}@illinois.edu}
}

\iclrfinalcopy %
\begin{document}

\maketitle

\begin{figure}[H]
\centering
\includegraphics[width=0.99\linewidth]{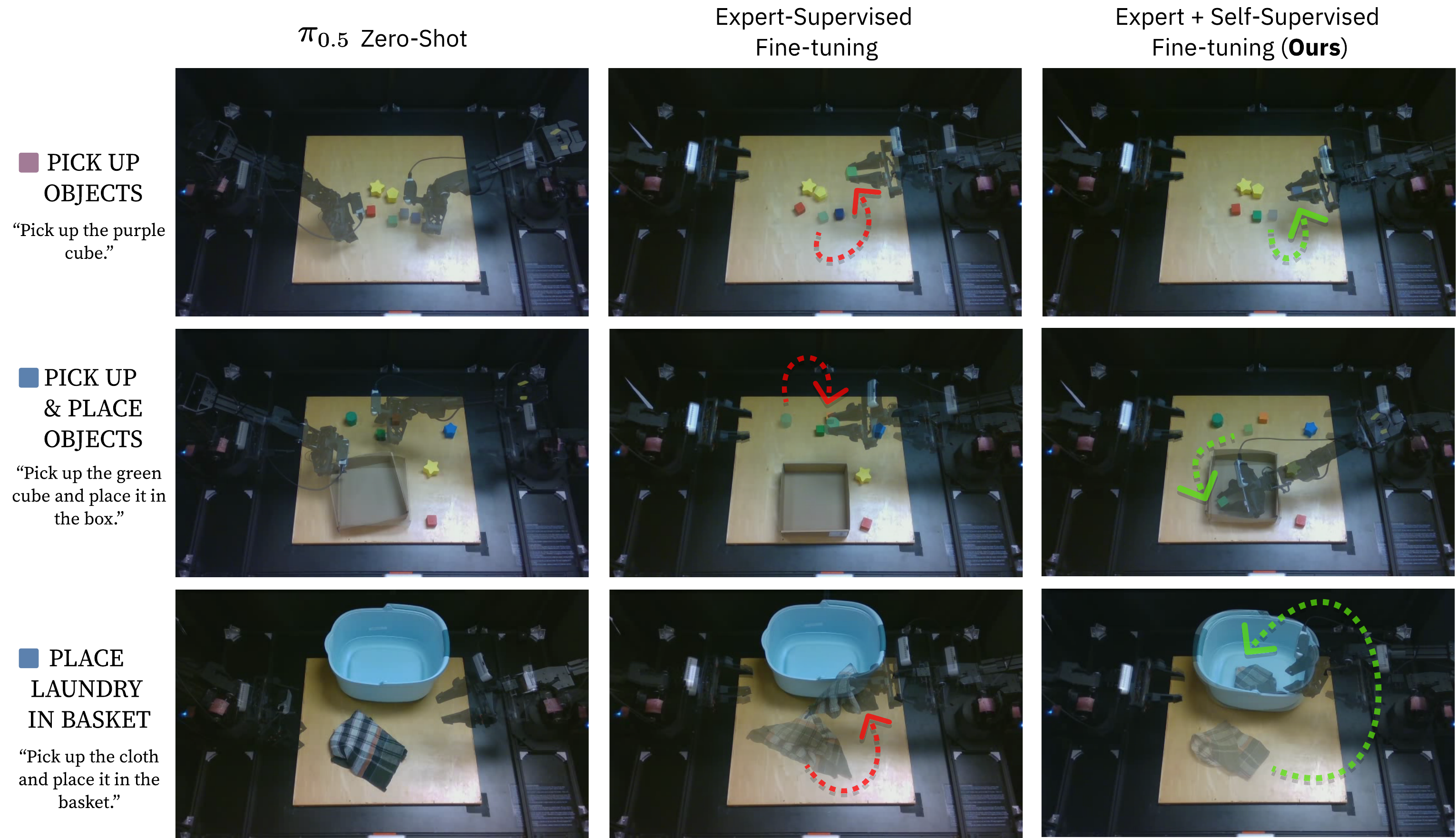}
    \caption{\textbf{Left}: Zero-shot $\pihalf$ policy follows instructions but cannot complete grasps on our robot. \textbf{Center}: Fine-tuning on expert ``pick-up'' data alone loses instruction following, reaching for the wrong object (rows 1--2), and loses the place motion after pick (rows 2--3). \textcolor{blue!40!black}{\textbf{Right (Ours)}: A single multi-task policy, jointly trained on 2.2 mins of \emph{expert-supervised} teleoperation for ``pick-up'' and \emph{self-supervised} generative rollouts from $\pihalf$ for ``pick-and-place" succeeds across all task families.}}
    \label{fig:demo-figure1}
\end{figure}

\begin{abstract}
State-of-the-art vision-language-action (VLA) models such as $\pihalf$ exhibit strong semantic understanding, instruction following and task behavior. However, when deployed on new robots, even minor mismatches in hardware configuration relative to pretraining can cause severe performance drops. Finetuning the VLA on in-domain expert data from the new embodiment improves performance on the expert task but leads to a loss in its original instruction following and behavioral priors. In this paper, we propose a self-supervised method that generates online interaction rollouts from the zero-shot VLA as additional training data for finetuning. Our experiments show this yields strong multi-task manipulation policies \textit{on the target robot} that (1) inherit pretrained tasks distilled from the zero-shot VLA without expert demos for them, (2) improve success and generalist instruction following beyond expert-only fine-tuning on expert-teleoperated tasks, and (3) mitigate forgetting of held-out tasks from the pretrained policy, not part of fine-tuning. We demonstrate the success of our approach across test sets probing generalization on a real ALOHA robot and a new simulation benchmark in RoboTwin. Video results are available at \url{https://self-supervised-control.pages.dev/}.
\end{abstract}

\section{Introduction}
Vision-language-action (VLA) policies \citep{genesisai2026gene, nvidiagear2025grootlatest, intelligence2025pi_05, intelligence2026pi_07, team2025gemini} learn to map a language instruction and camera images to robot actions, offering a promising route to task generalization in robotics. The catch is that this generalization is hard to inherit: a VLA policy trained on one robot platform and deployed on another of the same type can struggle with exact physical grounding and grasping due to minor variations in grippers, camera configurations or other embodiment gaps
\citep{song2026omniguide, wagenmaker2025steering}. Fine-tuning improves grasping behavior but erodes generalist instruction following and behavioral priors that made it valuable in the first place \citep{kachaev2025don, xu2025seeing, zha2026lap}. 
{\em How can we adapt a VLA to a new robot while retaining its instruction following ability and behavioral priors?}

When we deploy $\pihalf$~\citep{intelligence2025pi_05} to our ALOHA robotic platform~\citep{zhao2023learning} without tuning, the robot attempts to follow instructions, moving towards objects it should pick up, but fails to grasp them. For e.g., zero-shot policy localizes and attempts to ``pick up the purple cube" in Figure \ref{fig:demo-figure1}. If we fine-tune this VLA with ``expert-supervised'' teleoperated demonstrations from the ``pick-up objects" task family, this last mile fine-grained control is mostly restored, but instruction following degrades (Figure \ref{fig:demo-figure1}, center) as it picks up the green cube instead of the purple one. Additionally, expert supervision forgets place motion in the ``pick-and-place" tasks. 

Our key observation is that, even when it cannot complete a task, the base policy's action predictions stay semantically correlated with the instruction
across its online rollout trajectories (Figure \ref{fig:demo-figure1}, left).
We exploit this by adding ``self-supervised'' demonstrations. Concretely, we roll out the frozen base policy on a wider range of pick-and-place tasks similar to its pretraining, 
and train on its own predicted actions, a self-generated form of rehearsal that requires no access to the original pretraining data. Because these rollouts are executed on our target robot hardware, in our scenes, under our prompts, the rehearsal data carries no domain gap from the pretraining robot(s). Although the policy often fails to fully succeed on these rollouts, \added{it extends task coverage}, and training on them with a small number of expert supervised rollouts improves performance over expert-only fine-tuning across our tasks, including the pick-up task with expert supervision (Figure \ref{fig:demo-figure1}, right). 

We demonstrate the effectiveness of jointly training on expert-supervised and self-supervised demonstrations in both real and simulated environments. On the ALOHA robot, we evaluate expert tasks, self-supervised tasks, and held-out tasks; generalizing to novel objects and randomized object layouts throughout. We also contribute a new simulation benchmark and evaluation protocol in RoboTwin \citep{chen2025robotwin} that learns new tasks and retains old ones through self-demonstrations without access to the base policy's training data, testing generalization to novel objects and task compositions. Hence, we propose a more comprehensive way of evaluating generalization. Our \textbf{core insights} are that self-demos improve performance on (1) self-supervised tasks, (2) expert-supervised tasks, and (3) other held-out tasks not part of the fine-tuning data:
\begin{itemize}[leftmargin=*, nosep, itemsep=2pt]

\item \textit{Self-supervision distills behavioral priors of pretraining tasks without expert data for them.} Fine-tuning on expert ``pick" demos completely forgets ``place" behavior (0\% success rate). Jointly training on expert ``pick'' and self-supervised ``\added{pick-and-}place'' data recovers ``place'' behavior from 0\% to 55\% success
, without any expert ``place" demos. In Robotwin simulation, it recovers 10 severely forgotten mid-training tasks from 16.6\% to 70.6\%.

\item \textit{Training jointly on self- and expert-demos improves success
on expert tasks\replaced{.}{, including instruction following on out-of-distribution objects beyond expert-only fine-tuning.}} Success rate rises from 65\% to 90\% on the expert picking up cubes task, and from 40\% to 90\% on a \replaced{novel}{new} contact-rich bimanual gear insertion task (Figure~\ref{fig:DEMO_caterpillar}) \deleted{not present in pretraining} over expert-only fine-tuning. 
Moreover, sub-optimal self-demos are competent w.r.t. an oracle that re-collects expert demos for those same \added{``pick-and-place"} tasks, so there is little reason to pay the price of data re-collection \added{(Section \ref{sec:oracles})}. 

\item \textit{Self-demonstrations even preserve held-out skill families the policy was never \replaced{post-trained}{fine-tuned} on.} We conducted a user study for ``push'' and ``pour/turn over" rollouts on ALOHA robot, ours attempts the correct push motion in 82\% vs 38\% rollouts and pour/turn in 94\% vs 43\% compared to an oracle given expert teleoperated demos for all tasks (expert and self-supervised). Six blind raters preferred our motion in 68\% and 75\% of paired comparisons ($p<0.001$).

\end{itemize}

\label{sec:intro}

\section{Related Works}
\textbf{VLA post-training and continual learning.} Fine-tuning foundation VLMs on embodied action data frequently causes catastrophic forgetting, degrading semantic understanding and instruction following \citep{driess2026knowledge, gao2026taxonomy, hancock2025actions, kachaev2025don}. While recent VLA pre-training frameworks attempt to preserve these capabilities through multi-modal co-training \citep{wen2025dexvla, yang2025instructvla, driess2026knowledge, intelligence2025pi_05} or Bayesian factorization \citep{xu2025seeing}, downstream post-training to a target robotic domain remains prone to forgetting. Recent continual adaptation approaches explore visual representation alignment in simulation  \citep{kachaev2025don}, parameter expansion and adapter routing \citep{romer2026clare, wu2025continually}, weight merging \citep{yadav2025robust, fu2025mergevla}, and on-policy reinforcement fine-tuning \citep{hu2026simple, liu2026towards}. Yet, experience replay \citep{rolnick2019experience, chaudhry2019tiny}, which stores and rehearses past rollouts remains the most effective \citep{syed2025expres, liu2026vla, liu2026towards}. Existing rehearsal works assume access to original pre-training datasets \citep{kim2024openvla} or conduct intermediate fine-tuning on mid-scale benchmarks which completely overwrites pretrained representations \citep{liu2023libero, zhou2025libero}. In contrast, we address the practical setting where downstream users receive a pretrained VLA (e.g., $\pi_{0.5}$) without access to its proprietary pre-training data or task list. In a first, we propose distilling semantic priors and task behaviors directly from online rollouts of the zero-shot base policy during post-training, enabling seamless policy transfer to target user's setup.

In the continual learning literature, our method is similar to \textbf{generative replay} \citep{shin2017continual}, here applied to text-, vision-, and state- conditioned generative control; drawing a close parallel to prior-preserved fine-tuning of text-to-image diffusion models such as DreamBooth \citep{ruiz2023dreambooth, kumari2023multi, masip2025continual} and dataset distillation \citep{wang2018dataset}. \added{Unlike DreamBooth, which samples the pretrained model's prior \emph{offline}, we sample the base policy \emph{online and closed-loop} on the target robot, executing each action chunk before sampling the next. A second difference is that DreamBooth preserves the prior for the same class being customized (like dog), whereas we replay pretraining tasks disparate from those receiving expert supervision, to distill them. Both differences follow from generating the replay data through physical interaction on a target robot that may differ in hardware. Our approach is ``\textbf{self-supervised}'' because the base policy supplies both sides of a training pair, the states reached through its own execution on the target robot, and the action chunks generated which serve as pseudo-targets for distillation. The term ``\emph{self-supervision}'' here is different from its usage in classical robotics, where it refers to assigning rewards for self-improvement~\citep{pinto2016supersizing, oh2018self}. Unlike trajectory-based generative replay that trains a separate generator or teacher network to sample synthetic data~\citep{yue2024t}, we use the frozen pretrained VLA directly as its own trajectory generator to perform \emph{self-supervised self-distillation}.}

\label{sec:related}

\section{Method}
Our goal is to finetune a pre-trained VLA to solve a variety of tasks on our robot in our environments. VLAs do not perform well on different robots in a zero-shot manner and need to be finetuned on expert data from the specific deployment scenario. Existing finetuning methods are ineffective and also lead to forgetting. We propose a modification for how to finetune with expert data in section \ref{expert-supervision method} and also develop a new self-supervised methodology in section \ref{self-supervision method}.

\subsection{Preliminaries}
\label{sec:prelim}
Our base model is a pretrained continuous action chunking VLA policy $\pihalf$ \citep{intelligence2025pi_05}. For a trajectory with task prompt $p$ specified in natural language, the policy takes in a multi-modal observation $\mathbf{o}_t = (\mathbf{I}_t, p, \mathbf{q}_t)$ at each decision-making time step $t$, where $\mathbf{I}_t$ denotes camera images and $\mathbf{q}_t$ is the robot's proprioceptive state configuration. $\mathbf{I}_t$ and $\mathbf{q}_t$ vary with $t$ while $p$ is shared across the trajectory. The policy outputs a continuous action chunk $\mathbf{a}_{t} = a_{t:t+H}$ and models the probability distribution $\pi_{\theta}(\mathbf{a}_{t}|\mathbf{o}_t)$.

The policy architecture includes a specialized action expert and a VLM backbone.  The action expert
takes in a sequence of continuous noise action tokens, $\mathbf{a}_{t}^{\tau, \omega} = \tau \mathbf{a}_{t} + (1-\tau) \omega, \omega \sim \mathcal{N}(0, \mathbf{I})$, where $\tau \in [0, 1]$ is the flow matching time index and the action expert is trained to predict the flow velocity vector $\omega - \mathbf{a}_{t}$. The output action space is absolute joint angles $\mathbf{a}_{t} \in \mathbb{R}^{H \times D}$, with an action horizon $H$ and action dimension of $D$. The VLM backbone takes in image patch tokens encoded by the SigLIP vision encoder and a set of text tokens encoded through the PaliGemma \citep{beyer2024paligemma} vocabulary. The text tokens consist of an ordered sequence of prompt tokens $p$, discretized state $\mathbf{q}_t$, and an additional set of discrete action tokens called FAST $\mathbf{a}_{1:M}^{\text{FAST}}$. FAST tokens are only used during training and mapped from ground truth actions via a discrete cosine transform quantization \citep{pertsch2025fast}. 

\subsection{Expert-Supervised Imitation Learning}
\label{expert-supervision method}
\begin{figure*}[t]
\centering
\includegraphics[width=0.999\textwidth]{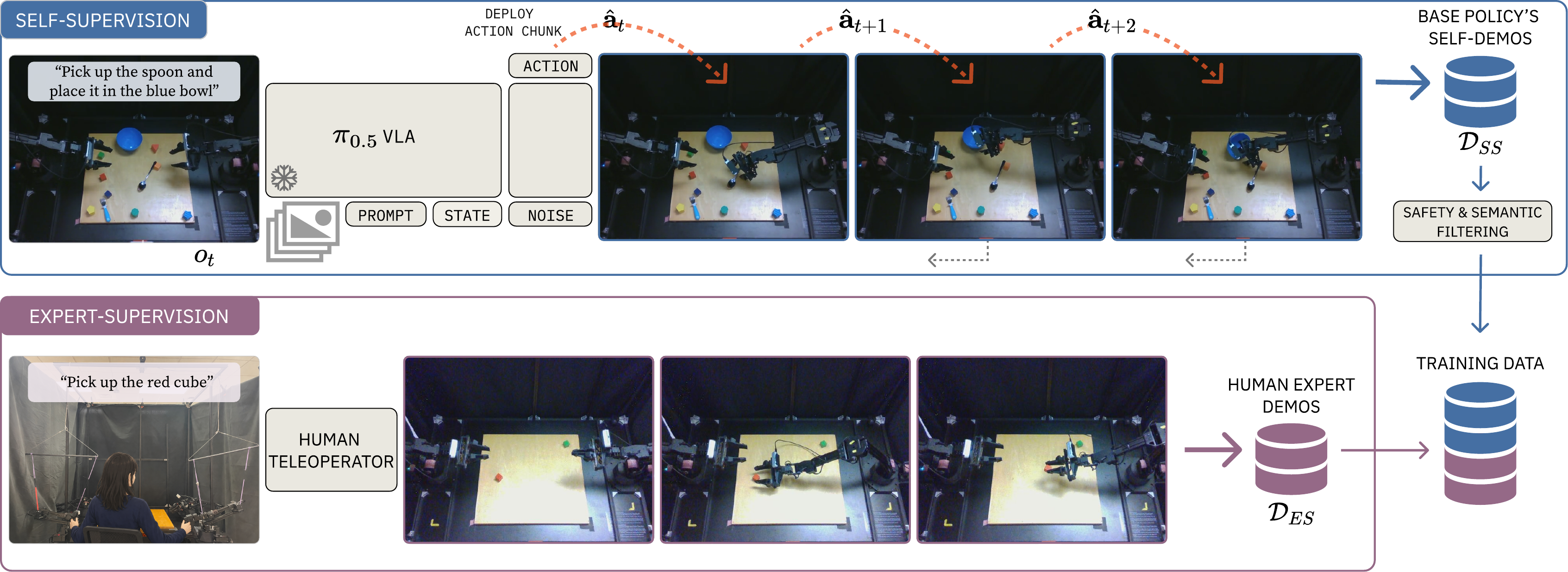} 
    \caption{\textbf{A new way to fine-tune pretrained VLAs.} We combine two multi-task data sources: (1) \emph{Bottom row:} expert-supervised demonstrations on a new task (E.g. ``Pick up the red cube"), and (2) \emph{Top row:} real-world interaction data generated by rolling out the frozen base policy on self-supervised tasks drawn from its pretraining distribution on the target robot (E.g. ``Pick up the spoon and place it in the blue bowl"). Rehearsing the expert demonstrations alongside the base policy's self-demonstrations preserves its semantic understanding, instruction following, and behavioral priors; yielding a multi-task policy that can be deployed to perform both new and pretrained tasks on the target robot. Red arrows ($\textcolor{red}{\dashrightarrow}$) denote action chunks generated by the base VLA policy.}
    \label{fig:main method}
\end{figure*}

We define a multi-task fine-tuning setup for imitation learning. Let $\mathcal{D}_{\text{ES}} = \{(\mathbf{o}_t, \mathbf{a}_{t})\}$, with $\mathbf{o}_t = (\mathbf{I}_t, p_{\text{ES}}, \mathbf{q}_t)$, be a dataset of observation--action pairs drawn from $N$ \emph{expert supervised} (ES) trajectories collected on our embodiment using teleoperation or a motion planner (Figure \ref{fig:main method} bottom row), where each trajectory is associated with a task prompt $p_{\text{ES}} \in \mathcal{P}_{\text{ES}}$. $\mathcal{P}_{\text{ES}}$ is selected such that the task behavior (which object to interact with and how) should be identifiable through vision and text prompts, and not just rely on vision. 

The model is optimized through a dual training objective, combining cross-entropy on the discretized FAST tokens and conditional flow matching on the continuous actions. For an observation $\mathbf{o}_t$ and target action chunk $\mathbf{a}$, we define the per-sample loss:
\begin{equation}
\ell(\mathbf{o}_t, \mathbf{a};\, \theta) = \mathbb{E}_{\tau,\, \omega} \Big[\, L_{\text{CE}}\big(\mathbf{a}_{1:M}^{\text{FAST}},\, f_\theta^{\ell}(\mathbf{o}_t, \mathbf{a}_{<m}^{\text{FAST}})\big) \;+\; \alpha \,\big\| (\omega - \mathbf{a}) - f_\theta^{a}(\mathbf{a}^{\tau, \omega}, \mathbf{o}_t) \big\|^2 \,\Big],
\label{eq:dual_objective}
\end{equation}
where $L_{\text{CE}}(\cdot, \cdot)$ is the cross-entropy loss applied only over the FAST action token positions and tends to train and converge the VLM backbone better on actions. $f_\theta^{a}$ is the action expert's predicted velocity, and $\alpha \in \mathbb{R}$ is a trade-off coefficient balancing the discrete and continuous objectives. The expert-supervised objective is the expectation of this loss over $\mathcal{D}_{\text{ES}}$:
\begin{equation}
\mathcal{L}_{\text{ES}}(\theta) = \mathbb{E}_{(\mathbf{o}_t, \mathbf{a}_t) \sim \mathcal{D}_{\text{ES}}} \big[\, \ell(\mathbf{o}_t, \mathbf{a}_t;\, \theta) \,\big].
\label{eq:es_loss}
\end{equation}

The original $\pihalf$ paper used both FAST actions and sub-task generation, but these two components are not released as part of the openpi code. We re-implement the FAST conditioning for training, as it tends to help generalization. We skip the sub-task generation head, as the goal of our work is to understand the properties and behaviors of the underlying vision-, text-, and state-conditioned flow matching policy. We find that the best performance on new tasks is achieved by fully fine-tuning both the VLM and the action expert and use that as default, with more ablations shown in Table~\ref{tab:robotwin_ablations}. 

\subsection{Self-Supervised Imitation Learning}
\label{self-supervision method}
In addition to the expert imitation data, we propose to collect the pretrained policy's self-demonstrations on familiar tasks. Concretely, we roll out the frozen base VLA checkpoint ${\pihalf}^{\text{base}}$ prompting it on a set of self-supervised task instructions $p_{\text{SS}} \in \mathcal{P}_{\text{SS}}$ that are within the base model's pretraining family of tasks, \emph{on our robot}. This generates trajectory motion behavior that correlates with the task prompt from the base policy's priors on our specific robot. At each decision step $t$ along the rollout, we sample a teacher action chunk from the base policy's conditional flow distribution,
\begin{equation}
\hat{\mathbf{a}}_{t} \sim {\pihalf}^{\text{base}}\big(\cdot \mid \mathbf{I}_t, p_{\text{SS}}, \mathbf{q}_t\big),
\label{eq:teacher_sampling}
\end{equation}
execute these actions to take steps in the environment, and record the resulting observation--action pair. Aggregating these pairs over all trajectory rollout steps and self-supervised tasks yields a rehearsal dataset $\mathcal{D}_{\text{SS}} = \{(\mathbf{o}_t, \hat{\mathbf{a}}_{t})\}$, with $\mathbf{o}_t = (\mathbf{I}_t, p_{\text{SS}}, \mathbf{q}_t)$ as shown in the top row of Figure \ref{fig:main method}. 
Since both observations and actions are collected along the teacher's own rollouts, every state is reached by executing the base policy's actions on our robot, in our scenes, under our prompts, so there is no embodiment or domain gap with respect to deployment. 
The self-supervised objective is the expectation of the per-sample loss in Eq.~\ref{eq:dual_objective} over $\mathcal{D}_{\text{SS}}$, with the teacher's sampled predictions $\hat{\mathbf{a}}_t$ (Eq.~\ref{eq:teacher_sampling}) serving as target actions:
\begin{equation}
\mathcal{L}_{\text{SS}}(\theta) = \mathbb{E}_{(\mathbf{o}_t, \hat{\mathbf{a}}_t) \sim \mathcal{D}_{\text{SS}}} \big[\, \ell(\mathbf{o}_t, \hat{\mathbf{a}}_t;\, \theta) \,\big].
\label{eq:ss_loss}
\end{equation}

The model is then jointly optimized on both objectives:
\begin{equation}
\mathcal{L}(\theta) = \mathcal{L}_{\text{ES}}(\theta) + \lambda \, \mathcal{L}_{\text{SS}}(\theta),
\label{eq:joint_loss}
\end{equation}
where $\lambda \in \mathbb{R}$ controls the relative weight of the self-supervision term. Interestingly, despite often unsuccessful in completing the task, the self-supervised demonstrations improve learning by increasing task coverage with the new robot/environment and retaining instruction following and pretrained behavior through distillation.

\textbf{Self-demo filtering.} We conduct semantic and safety filtering on the self-demos. A rollout is discarded if it never approaches the specified object even once (Figure \ref{fig:selfdemo_timeline} shows self-demo distribution after filtering). We also filter out any self-demos with undesired behavior before distillation, e.g., if it makes unsafe contact such as wedging the grippers against the basket's rim in laundry task or the arm and eye-in-hand camera going inside the basket and getting stuck. 

\textbf{Data collection and mixture.} We collect trajectories with different task instructions from the same initial scene, to ensure the model continues to use the text instruction rather than only the visual observation to identify the task. Training with a balance of expert- and self-supervised data performs best, using a 1:1 mixture and sampling uniformly in each. Task-weighted sampling hurts performance. In experiments with a physical robot, training with strongly imbalanced ratios leads to artifacts and erratic behavior.

\label{sec:method}

\section{Proposed Benchmarks and Protocol}

\begin{figure}[t]
\centering
\includegraphics[width=0.999\linewidth]{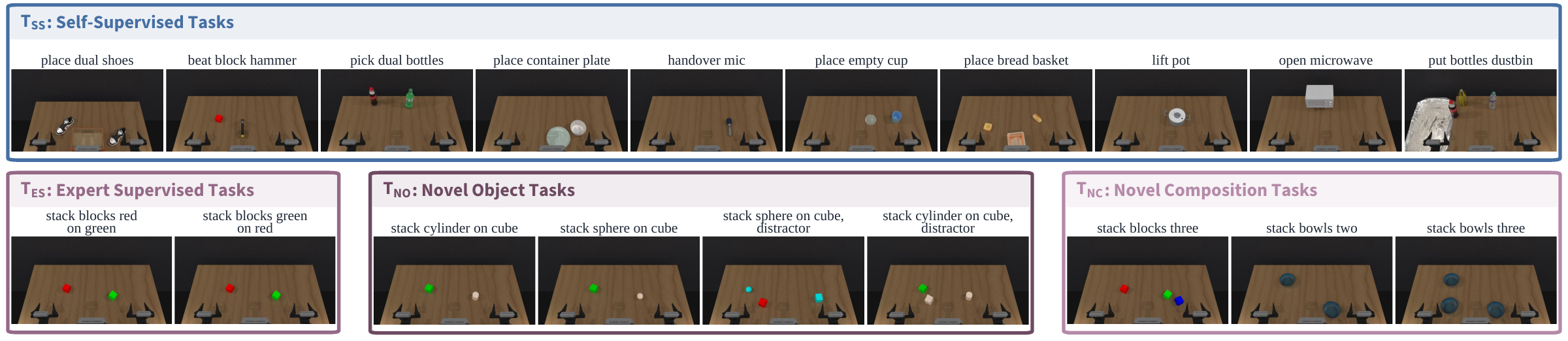}
    \caption{\textbf{Our simulation benchmark in RoboTwin} is designed to test generalization to new object layouts in task families that are (1) expert-supervised, $T_{ES}$ (stacking two blocks) and (2) self-supervised, $T_{SS}$ (broader set of pick-place, lifting, opening tasks). We also test generalization to (3) novel objects within expert tasks, $T_{NO}$ and (4) new compositions of expert skills, $T_{NC}$.}
    \label{fig:task-overview-robotwin}
\end{figure}

We experiment on bimanual robots: the stationary \textbf{ALOHA-1} \citep{zhao2023learning} in the real world and the ALOHA-Agilex embodiment in \textbf{RoboTwin 2.0} \citep{chen2025robotwin} in simulation. In real we post-train the pretrained $\pihalf$ directly; in sim we mid-train first, then post-train. As summarized in Table \ref{tab:data_splits}, our benchmarks evaluate generalization across four test distributions: $\mathbf{T_{ES}}$: tasks from the distribution of expert supervised training; $\mathbf{T_{SS}}$: tasks from the distribution of self-supervised training; $\mathbf{T_{NO}}$: tasks with similar skills to $T_{ES}$ but with novel objects not seen during training; and $\mathbf{T_{NC}}$: new tasks composed of skills from $T_{ES}$, e.g. stacking three blocks instead of two (RoboTwin only, see Figure \ref{fig:task-overview-robotwin}). All our test sets use random and diverse initial object positions. 

\added{In simulation, most experiments post-train in the same embodiment used for mid-training (\textbf{E1}), isolating forgetting caused by fine-tuning alone. We additionally introduce a controlled embodiment gap by changing the head camera view (\textbf{E2}; Figure \ref{fig:robotwin-embodiments}, Table \ref{tab:robotwin_embodiment_gap_result}), measuring how this gap compounds forgetting. While simulation gives systematic, repeatable evaluation; real robot experiments reflect the practical regime that motivates our work: no mid-scale fine-tuning dataset exists for most specific robots,
hence, small, embodiment-specific expert datasets must suffice for post-training. It also exposes the VLA's zero-shot instruction following on the robot since it was pretrained only on real data, and an embodiment gap that prevents zero-shot grasps.}

\label{sec:benchmark}

\section{Experiments}

\textbf{Implementation Details.}
The ALOHA has 7 DoFs in each arm (6 target joint angles and 1 parallel-jaw gripper position). $\pihalf$ predicts action chunks $\mathbf{a}_{t} \in \mathbb{R}^{H \times D}$, with $H=50, D=32$. In both real and sim, $H/2=25$ actions are executed open loop on the robot before querying the VLA for the next observation from the environment. All real experiments are at a control frequency of 30-50 Hz.

\textbf{Metrics.} \emph{\textbf{Success rate (SR)}} measures if the policy interacts with the specified object, demonstrates the correct behavior, and completes the task. A policy may largely do the right thing, exhibit task relevant behavior but ultimately fail leading to a low success rate. Hence, we also report \emph{\textbf{Instruction Following (IF)}}: \emph{\textbf{IF (pre-grasp)}}, measuring how often the robot approaches and attempts to grasp the specified object. For multi-stage tasks, we additionally report \emph{\textbf{IF (post-grasp)}}, i.e. how often the arm moves to the correct container or insertion peg after an attempted pick-up. 

\subsection{Zero-shot VLA behavior}

\smallskip\noindent\textit{Real.} \textbf{Base policy follows instructions zero-shot but cannot grasp on our robot (embodiment gap).} For Benchmark 1 in Table \ref{tab:real_robot_results}, \emph{$\pihalf$ zero-shot} localizes and reaches the prompted object 80\% on picking red/green cubes, 70\% on picking novel objects but \emph{grasps none} (0\% SR). We attribute this to potential gaps in exact hardware configuration between our target robot and the pretraining data collection system, such as grippers used and top camera's pose. On the self-supervised pick-and-place tasks, it exhibits pretraining behavior: it moves to pick correct object 70\% and laundry 65\% times, fails grasp, moves towards container ``Y" 25\% and laundry basket 30\% attempting to place, but always fails (0\% SR; the 5\% SR on laundry is an accidental gripper slip). 

\smallskip\noindent\textit{Sim.} \textbf{RoboTwin Base Policy.} \emph{$\pihalf$ zero-shot} is not reported to contain simulation data in its pretraining \citep{intelligence2025pi_05}, scoring 0\% on all tasks. We therefore mid-train on these 10 tasks with expert supervision from RoboTwin's motion planner to obtain a \emph{RoboTwin base policy} (90.8\%, Table \ref{tab:robotwin_main_ES_result}). Unlike prior benchmarks that assume persistent access to mid-training data, we restrict access to it during post-training, mirroring real-world deployment. \added{Under the E2 embodiment gap, this base policy drops to 51.6\% on $T_{SS}$ zero-shot (Table~\ref{tab:robotwin_embodiment_gap_result}).}

\subsection{Catastrophic forgetting in VLA post-training}

\smallskip\noindent\textit{Real.} \textbf{Fine-tuning to close embodiment gap causes forgetting of generalist instruction following; multi-task better than single-task.}
Fine-tuning on 30 expert demos of a single task, ``Pick up the green cube'' (\emph{Single Task ES}) degrades instruction following, relying more on visual observations collapsing the text conditioning. This ignorance of text has been called `visual-action shortcut learning' in \citet{xu2025seeing}. Splitting the same budget across two tasks in a shared scene with both objects in each scene (15 ``red'' + 15 ``green'') recovers it: \emph{Multi-Task ES} improves IF from 40\% to 50\% on $T_{NO}$ and 45\% to 75\% on $T_{SS}$. We also ablate this choice of multiple instructions in the same scene (requires text to specify the task) by training \emph{Multi-Task ES, Single Object/Scene} with 15 ``red'' and 15 ``green'' demos having only one object in each scene and find it has worse IF.

\smallskip\noindent\textit{Real.} \noindent \textbf{Expert supervision alone forgets pretraining task motion priors.}
Despite better instruction following, \emph{Multi-Task ES} overfits the motion to the expert task family (``Pick up X'') at the expense of the pretrained pick-and-place behavior 
\replaced{(post-grasp IF for placing drops from 25\% to 0\%).}{(25\% and 30\% zero-shot post-grasp IF for placing)}.

\begin{table}[t]
\centering
\caption{\textbf{ALOHA Robot.} Multi-task policy results across a variety of \emph{pick up-only} and \emph{pick-and-place} tasks. Our method not only outperforms $\pihalf$ zero-shot and fine-tuning baselines for tasks that are expert demonstrated $T_{ES}$ like ``Pick up red/green cube'', but also enables task success on various pick and place tasks purely distilled through the zero-shot policy's self-supervised demos. Our approach enables fine-tuning to get a MT policy ready to deploy on our robot with only 2.2 minutes of expert teleoperation data (30 episodes, 6121 frames), while retaining strong generalization priors from the base VLA. Each row's results are from a single model. All values are percentages over 20 trials, except \emph{Single Task ES} on $T_{ES}$ which uses 10 trials.}
\label{tab:real_robot_results}
\renewcommand{\arraystretch}{1.15}
\resizebox{\textwidth}{!}{%
\begin{tabular}{@{}l @{\hspace{1.0em}} *{2}{c} | *{2}{c} | *{3}{c} | *{3}{c}@{}}
\toprule
& \multicolumn{2}{c|}{\textbf{$T_{ES}$: Expert Supervised Tasks}} & \multicolumn{2}{c|}{\textbf{$T_{NO}$: Novel Object Tasks}} & \multicolumn{6}{c}{\textbf{$T_{SS}$ Self-Supervised Tasks}} \\
& \multicolumn{2}{c|}{\textit{(``Pick up red/green cube'')}} & \multicolumn{2}{c|}{\textit{(``Pick up X'')}} & \multicolumn{3}{c|}{\textit{(``Pick up X and place it in Y'')}} & \multicolumn{3}{c}{\textit{(``Pick up laundry and place in basket'')}} \\
\cmidrule(lr){2-3} \cmidrule(lr){4-5} \cmidrule(lr){6-8} \cmidrule(lr){9-11}
\textbf{Method}
& IF (pre-grasp) & SR
& IF (pre-grasp) & SR
& IF (pre-grasp) & IF (post-grasp) & SR
& IF (pre-grasp) & IF (post-grasp) & Partial SR \\
\midrule
$\pihalf$ Zero-Shot               & 80 & 0  & \textbf{70} & 0  & 70 & 25  & 0  & 65 & 30  & 5 \\
\midrule
$\pihalf$ Single Task ES                       &  90 &  50 &  40 &  25 & 45 & 25 & 0 & — & — & — \\
$\pihalf$ Multi-Task ES, Single Object/Scene                       &  65 &  65 &  45 &  30 & 55 & 35 & 10 & — & — & — \\
\midrule
$\pihalf$ Multi-Task ES                         & \textbf{100} & 65 & 50 & 50 & 75 & 0  & 0  & \textbf{100} & 50 & 10 \\
\rowcolor{blue!8}
\textbf{$\pihalf$ Multi-Task ES+SS (Ours)} & \textbf{100} & \textbf{90} & 55 & \textbf{55} & \textbf{90} & \textbf{90} & \textbf{55} & \textbf{100} & \textbf{95} & \textbf{40} \\
\bottomrule
\end{tabular}%
}
\end{table}

\providecommand{\rh}[1]{\rotatebox[origin=c]{60}{#1}}
\begin{table}[t]
\centering
\renewcommand{\arraystretch}{1.15}
\newcommand{\subhead}[1]{\multicolumn{9}{@{}l}{\makebox[0pt][l]{\textbf{\textit{#1}}}}}
\setlength{\tabcolsep}{3pt}
\caption{\textbf{RoboTwin Simulation (E1: default embodiment).} In mid-training, policies are fine-tuned on 10 tasks. Post-training on new expert tasks ($T_{ES}$) while rehearsing self-demos from mid-training stage ($T_{SS}$) mitigates catastrophic forgetting in expert-only. Success rates (\%, $\uparrow$) averaged across 50 seeds per task.}
\label{tab:robotwin_main_ES_result}
\resizebox{0.98\textwidth}{!}{%
\begin{tabular}{@{} l | *{10}{c} | c | cc | c @{}}
\toprule
& \multicolumn{11}{c|}{\textbf{$T_{SS}$: Mid-training tasks (old)}} & \multicolumn{3}{c}{\textbf{$T_{ES}$: Post-training tasks (new)}} \\
\cmidrule(lr){2-12} \cmidrule(lr){13-15}
\textbf{Experiment}
& \rh{Place Shoes} & \rh{Beat Hammer} & \rh{Pick Bottles} & \rh{Place Plate} & \rh{Handover Mic}
& \rh{Place Cup} & \rh{Place Basket} & \rh{Lift Pot} & \rh{Open Micro.} & \rh{Put Dustbin}
& \rh{\textbf{Avg. $T_{SS}$}} & \rh{Stack $R \rightarrow G$} & \rh{Stack $G \rightarrow R$} & \rh{\textbf{Avg. $T_{ES}$}} \\
\midrule
$\pi_{0.5}$ Zero-Shot            & -- & -- & -- & -- & -- & -- & -- & -- & -- & -- & 0 & -- & -- & -- \\
\subhead{Mid-Training Experiments} \\[0.5pt]
$\pi_{0.5}$ RoboTwin Base Policy & 76 & 94 & 98 & 100 & 100 & 98 & 82 & 98 & 72 & 90 & \textbf{90.8} & -- & -- & -- \\
\midrule
\subhead{Post-Training Experiments: All models initialized from RoboTwin Base Policy} \\[0.5pt]
Multi-Task ES                    & 0 & 0 & 0 & 60 & 0 & 36 & 34 & 36 & 0 & 0 & 16.6 & 90 & 96 & 93 \\
\rowcolor{blue!8}
\textbf{Multi-Task ES+SS (Ours)} & \textbf{42} & \textbf{82} & \textbf{48} & \textbf{92} & \textbf{62} & \textbf{92} & \textbf{58} & \textbf{86} & \textbf{66} & \textbf{78} & \textbf{70.6} & \textbf{98} & \textbf{98} & \textbf{98} \\
\bottomrule
\end{tabular}%
}
\end{table}

\begin{table}[b]
\centering
\renewcommand{\arraystretch}{1.15}
\newcommand{\subhead}[1]{\multicolumn{9}{@{}l}{\makebox[0pt][l]{\textbf{\textit{#1}}}}}
\setlength{\tabcolsep}{3pt}
\caption{\added{\textbf{RoboTwin under an embodiment gap (E2).} We change only the head camera to point straight down. Zero-shot, the RoboTwin Base Policy (mid-trained in the default embodiment, E1) drops from 90.8\% to 51.6\% on $T_{SS}$. Post-training on new expert tasks ($T_{ES}$) in E2 while rehearsing self-demos collected in E2 on mid-training tasks $T_{SS}$ mitigates forgetting from 21.4\% to 36.0\%. Success rates (\%, $\uparrow$) averaged across 50 seeds per task.}}
\label{tab:robotwin_embodiment_gap_result}
\resizebox{0.99\textwidth}{!}{%
\begin{tabular}{@{} l | *{10}{c} | c | cc | c @{}}
\toprule
& \multicolumn{11}{c|}{\textbf{$T_{SS}$: Mid-training tasks (old)}} & \multicolumn{3}{c}{\textbf{$T_{ES}$: Post-training tasks (new)}} \\
\cmidrule(lr){2-12} \cmidrule(lr){13-15}
\textbf{Experiment}
& \rh{Place Shoes} & \rh{Beat Hammer} & \rh{Pick Bottles} & \rh{Place Plate} & \rh{Handover Mic}
& \rh{Place Cup} & \rh{Place Basket} & \rh{Lift Pot} & \rh{Open Micro.} & \rh{Put Dustbin}
& \rh{\textbf{Avg. $T_{SS}$}} & \rh{Stack $R \rightarrow G$} & \rh{Stack $G \rightarrow R$} & \rh{\textbf{Avg. $T_{ES}$}} \\
\midrule
\subhead{Mid-Training Experiments (E1: Default embodiment)} \\[0.5pt]
$\pi_{0.5}$ RoboTwin Base Policy (E1)         & 76 & 94 & 98 & 100 & 100 & 98 & 82 & 98 & 72 & 90 & \textbf{90.8} & -- & -- & -- \\
\midrule
\subhead{Embodiment Gap Experiments (E2: Topview embodiment)} \\[0.5pt]
$\pi_{0.5}$ RoboTwin Base Policy Zero-Shot (E1 $\rightarrow$ E2) & 6 & 80 & 64 & 90 & 68 & 24 & 24 & 48 & 62 & 50 & \textbf{51.6} & 4 & 8 & 6 \\
\midrule
\subhead{Post-Training Experiments in E2: All models initialized from RoboTwin Base Policy} \\[0.5pt]
Multi-Task ES (Topview)                       & 0 & 2 & 0 & 62 & 0 & \textbf{46} & \textbf{38} & \textbf{50} & 16 & 0 & 21.4 & \textbf{96} & 94 & \textbf{95} \\
\rowcolor{blue!8}
\textbf{Multi-Task ES+SS (Topview, Ours)}     & \textbf{2} & \textbf{60} & \textbf{62} & \textbf{70} & \textbf{44} & 24 & 30 & 32 & \textbf{18} & \textbf{18} & \textbf{36.0} & 92 & \textbf{96} & 94 \\
\bottomrule
\end{tabular}%
}
\end{table}

\smallskip\noindent\textit{Sim, E1.} \textbf{Multi-Task expert fine-tuning forgets catastrophically.}
Similar to our real benchmark, expert post-training is done on two stacking tasks in a shared scene (``Stack red block on green block" and ``Stack green block on red block"). Naive \emph{Multi-Task ES} fine-tuning learns $T_{ES}$ well (93\%), but catastrophically forgets mid-training tasks \replaced{$T_{SS}$ from 90.8\% to 16.6\%, with six tasks falling to 0\%, Table \ref{tab:robotwin_main_ES_result}.}{($T_{SS}$), dropping by 74.2\% on average.}

\added{\smallskip\noindent\textit{Sim, E2.} \textbf{The same forgetting occurs under an embodiment gap.} In E2, \emph{Multi-Task ES} again learns $T_{ES}$ (95\%), while $T_{SS}$ falls from 51.6\% to 21.4\%, with five tasks at 0--2\% (Table~\ref{tab:robotwin_embodiment_gap_result}).}

\subsection{Rehearsing self-demos improves both self-supervised and expert tasks}

\smallskip\noindent\textit{Real.} \textbf{Benchmark 1: pick-up and pick-and-place.} \emph{Multi-Task ES+SS} (i) distills the place behavior \emph{purely} from the base policy's self-demos, with no expert place demos, reaching 90\% post-grasp IF and 55\% SR on pick-and-place as compared to 0\% for Multi-Task ES; (ii) closes the hardware embodiment gap through expert data, reaching 90\% SR on the expert cubes and 55\% on novel objects; (iii) attains the best IF on $T_{ES}$ and $T_{SS}$ test sets; achieving the best SR on $T_{NO}$, grasping every object it localizes. 
(iv) Achieves 40\% partial SR on the harder laundry task where self-demo priors are noisy and the cloth needs to be lifted over the baskets rim to place.  

\begin{table}[h]
\centering
\caption{\textbf{ALOHA Robot Caterpillar.} This is a complex, long-horizon bimanual task of colored peg insertion on a non-stationary board. This task is out-of-distribution w.r.t. base policy's pretraining. Using expert- and self-supervised fine-tuning (\emph{Multi-task ES+SS}) improves all metrics across in-distribution gear colors ($T_{ES}$) and out-of-distribution gear colors ($T_{NO}$). (\%, 10 trials)}
\label{tab:caterpillar}
\renewcommand{\arraystretch}{1.15}
\resizebox{0.8\textwidth}{!}{%
\begin{tabular}{@{}l @{\hspace{1.0em}} *{3}{c} | *{3}{c}@{}}
\toprule
& \multicolumn{3}{c|}{\textbf{$T_{ES}$: Expert Supervised Tasks}} & \multicolumn{3}{c}{\textbf{$T_{NO}$: Novel Object Tasks}} \\
& \multicolumn{3}{c|}{\textit{(``Pick up gear X and insert it into gear shaft X'')}} & \multicolumn{3}{c}{\textit{(``Pick up gear X and insert it into gear shaft X'', novel gears)}} \\
\cmidrule(lr){2-4} \cmidrule(lr){5-7}
\textbf{Method}
& IF (pre-grasp) & IF (post-grasp) & SR
& IF (pre-grasp) & IF (post-grasp) & SR \\
\midrule
$\pihalf$ Zero-Shot                                  & 0 & 0 & 0 & 20 & 0 & 0 \\
\midrule
$\pihalf$ Multi-Task ES                      & 100\% & 40\% & 40\% & 70\% & 30\% & 10\% \\
\rowcolor{blue!8}
\textbf{$\pihalf$ Multi-Task ES+SS (Ours)} & \textbf{100\%} & \textbf{90\%} & \textbf{90\%} & \textbf{80\%} & \textbf{40\%} & \textbf{20\%} \\
\bottomrule
\end{tabular}%
}
\end{table}

\begin{figure}[h]
\centering
\includegraphics[width=\linewidth]{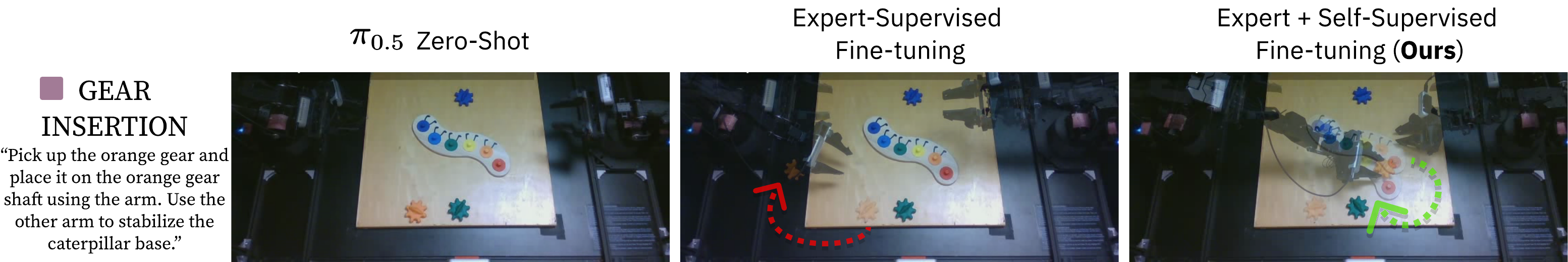}
    \caption{A rollout from the caterpillar test set showing successful alignment of orange gear on peg.}
    \label{fig:DEMO_caterpillar}
\end{figure}

\smallskip\noindent\textit{Real.} \textbf{Benchmark 2: caterpillar gear insertion.} Caterpillar gear insertion (Table~\ref{tab:caterpillar}) is a precise, contact-rich bimanual task: the policy must localize the specified gear, grasp it, move to the correct gear color, and align it onto the specified peg shaft. It is a peg-in-hole insertion with tight alignment tolerances, made harder by a non-fixed peg board (the other arm adjusts the board if required). We even vary board positions across test sets. The 0\% and 30\% \emph{$\pihalf$ zero-shot} instruction following indicates the task and its objects lie largely outside the base policy's pretraining. Adding self-supervision raises SR from 40\% to 90\% on the expert-demonstrated gear colors ($T_{ES}$), with post-grasp IF rising from 40\% to 90\%, Figure \ref{fig:DEMO_caterpillar}. On three unseen colors ($T_{NO}$), SR improves from 10\% to 20\% and post-grasp IF from 30\% to 40\%.

\smallskip\noindent\textit{Sim, E1.} \textbf{RoboTwin.} 
Rehearsing with only 10 self-demos per task in the simulator \emph{Multi-Task ES+SS} retains 70.6\% on $T_{SS}$ (78\% of the base policy's 90.8\%) versus 16.6\% for \emph{Multi-Task ES}.
We use the first 10 generated self-demos w/o filtering and recovery spans diverse skills (hammering, opening microwaves, pick-and-place, etc). It also achieves peak plasticity on expert stacking tasks (98\%) against 93\% with \emph{Multi-Task ES}.

\added{\smallskip\noindent\textit{Sim, E2.} \textbf{RoboTwin, embodiment gap.} Rehearsing 10 self-demos per task collected in E2, \emph{Multi-Task ES+SS} retains 36.0\% on $T_{SS}$ (70\% of the zero-shot 51.6\%) versus 21.4\% for \emph{Multi-Task ES}, with comparable $T_{ES}$ (94\% vs.\ 95\%). Gains are largest on beat block hammer, pick dual bottles, and handover mic.} 

\subsection{Comparison with upper bounds}
\label{sec:oracles}

\begin{table}[h]
\centering
\caption{\textbf{ALOHA Robot: comparison with full expert supervision.} \emph{ES+ES} is an upper-bound reference that collects expert human teleoperation on \textit{our} robot for \emph{all} tasks, both $T_{ES}$ and $T_{SS}$, replacing our self-supervised demonstrations with expert data on the same scenes and task list. \emph{Multi-Task ES+SS (Ours)} is the same checkpoint from Table \ref{tab:real_robot_results}. (\%, 20 trials)}
\label{tab:full-expert-supervision}
\renewcommand{\arraystretch}{1.15}
\resizebox{\textwidth}{!}{%
\begin{tabular}{@{}l @{\hspace{1.0em}} *{2}{c} | *{2}{c} | *{3}{c} | *{3}{c}@{}}
\toprule
& \multicolumn{2}{c|}{\textbf{$T_{ES}$: Expert Supervised}} & \multicolumn{2}{c|}{\textbf{$T_{NO}$: Novel Object}} & \multicolumn{6}{c}{\textbf{$T_{SS}$: Self-Supervised Tasks}} \\
& \multicolumn{2}{c|}{\textit{(``Pick up red/green cube'')}} & \multicolumn{2}{c|}{\textit{(``Pick up X'')}} & \multicolumn{3}{c|}{\textit{(``Pick up X and place it in Y'')}} & \multicolumn{3}{c}{\textit{(``Pick up laundry and place in basket'')}} \\
\cmidrule(lr){2-3} \cmidrule(lr){4-5} \cmidrule(lr){6-8} \cmidrule(lr){9-11}
\textbf{Method}
& IF (pre-grasp) & SR
& IF (pre-grasp) & SR
& IF (pre-grasp) & IF (post-grasp) & SR
& IF (pre-grasp) & IF (post-grasp) & Partial SR \\
\midrule
\rowcolor{gray!15}
$\pihalf$ Multi-Task ES+ES (Recollect Expert Oracle)         & 100 & 95 & 65 & 60 & 90 & 90 & 65 & 100 & 100 & 90 \\
\rowcolor{blue!8}
\textbf{$\pihalf$ Multi-Task ES+SS (Ours)}     & 100 & 90 & 55 & 55 & 90 & 90 & 55 & 100 & 95 & 40 \\
\bottomrule
\end{tabular}%
}
\end{table}

\smallskip\noindent\textit{Real.} 
The naive alternative to our method is to teleoperate expert demos for all tasks, including those in $T_{SS}$. This \emph{MT ES+ES} setting removes the data-availability constraint our method is designed for, and serves as an upper-bound reference point. The upper bound matches or slightly exceeds ours, Table~\ref{tab:full-expert-supervision}: 95\% vs. 90\% SR on $T_{ES}$, 65\% vs. 55\% on $T_{SS}$ pick-and-place, with instruction following tied on 3/4 task families. The gap is larger on the laundry task (90\% vs. 40\% SR). As self-demos are noisier, our \emph{MT ES+SS} policy releases the cloth near the edge of the basket and it often falls out.

\begin{wraptable}{r}{0.53\textwidth}
\vspace{-\baselineskip}
\centering\small
\renewcommand{\arraystretch}{1.1}\setlength{\tabcolsep}{3pt}
\caption{\textbf{RoboTwin Simulation.} Oracle baselines rehearse stored expert data from mid-training. \emph{Ours:} rehearsing 10 self-demos per task recovers 82.5\% of the rehearsal oracle's $T_{SS}$ performance, landing 0.2 points below it Overall and above Rehearsal + LoRA. Average test set SR (\%, $\uparrow$).}
\label{tab:robotwin_rehearsal}
\resizebox{\linewidth}{!}{%
\begin{tabular}{@{}l cccc c@{}}
\toprule
\textbf{Experiment} & $T_{SS}$ (old) & $T_{ES}$ (new) & $T_{NO}$ & $T_{NC}$ & \textbf{Overall} \\
\midrule
\rowcolor{gray!15}
Multi-Task ES+ES (Rehearsal Oracle)          & 85.6 & 97.0 & 30.0 & 15.3 & 57.0 \\
\rowcolor{gray!15}
Multi-Task ES+ES (Rehearsal + LoRA Oracle)   & 84.6 & 77.0 & 29.5 & 11.3 & 50.6 \\
\midrule
Multi-Task ES                                & 16.6 & 93.0 & \textbf{57.0} & 6.7  & 43.3 \\
\rowcolor{blue!8}
\textbf{Multi-Task ES+SS (Ours)}             & \textbf{70.6} & \textbf{98.0} & 44.5 & \textbf{14.0} & \textbf{56.8} \\
\bottomrule
\end{tabular}%
}
\end{wraptable}

\smallskip\noindent\textit{Sim.} \textbf{Oracle baselines} (Table~\ref{tab:robotwin_rehearsal}). These assume access to stored mid-train expert data, which our setup avoids. \emph{Rehearsal (Oracle)} fully fine-tunes on stored mid-train data plus new expert data. \emph{Rehearsal + LoRA (Oracle)} tests the claim from \citet{liu2026vla} that PEFT with replay mitigates forgetting. Ours instead generates 10 self-demos per task, trains on this mix of success and failure rollouts, without storing any old data.

\subsection{Self-demos retain held-out skill families which expert oracle forgets}
\label{sec:push discussion}

\smallskip\noindent\textit{Real.} In Table \ref{tab:user_study}, we evaluate the \emph{MT ES+ES} versus \emph{MT ES+SS (Ours)} checkpoints from Table \ref{tab:full-expert-supervision} on two held-out skill families that are absent from both expert- or self-supervised data. We find self-supervision is more effective at retaining the motion priors of both ``Push objects'' and ``Pour/turn over'' skills. Since the presence of a motion prior in a rollout is subjective, we assess it with a blind user study rather than a fixed success criterion.
While IF is comparable across the 2 policies, motion prior is far better retained by our method. Ours attempts the requested push on 82\% of rated rollouts vs.\ 38\% for \emph{Multi-Task ES+ES}, and the pour/turn on 94\% vs.\ 43\% ($p<0.001$). On push, the oracle instead picks the container up in 43\% of rollouts.

\section{Ablation and Analysis}
\label{sec:ablation-analysis}
\textbf{Augmenting self-supervision helps even when expert data already covers the task.} Table~\ref{tab:es_vs_esss_pnp} isolates the contribution of augmenting pick-place self-demos in addition to expert pick-place teleoperation demonstrations for fine-tuning only on pick-place task family. Here, we conducted blind A/B testing of the two policies on all 20 scenes from $T_{SS}$ test set. 

\textbf{Qualitative Results.} Overall, our method yields smooth rollouts personalized to our target robot's desired motion range, as compared to the jerkier rollouts from the base policy which often raises the two arms higher than the task requires. 
Our method post-trains the VLA to stay sensitive to text instructions (which object to go to and how to interact with it), and has better recovery behavior when objects sit at table edges; it turns to look for objects behind the wrist camera unlike baselines.

\begin{wraptable}{r}{0.53\textwidth}
\vspace{-\baselineskip}
\centering\small
\renewcommand{\arraystretch}{1.1}\setlength{\tabcolsep}{3pt}
\caption{\added{\textbf{Fine-tuning ablations (RoboTwin).} We ablate PEFT and flow-only training loss for \emph{Multi-Task ES}. Full fine-tuning with FAST+Flow objective has best stability-plasticity trade-off while self-demos on top of that further mitigate forgetting.}}
\label{tab:robotwin_ft_main}
\resizebox{\linewidth}{!}{%
\begin{tabular}{@{}l cccc c@{}}
\toprule
\textbf{Experiment} & $T_{SS}$ & $T_{ES}$ & $T_{NO}$ & $T_{NC}$ & \textbf{Overall} \\
\midrule
Multi-Task ES (LoRA)             & 28.0 & 87.0 & 37.5 & 2.0 & 38.6 \\
Multi-Task ES (Flow-only)        & 11.2 & 94.0 & \textbf{57.0} & 5.3 & 41.9 \\
Multi-Task ES$^\dagger$          & 16.6 & 93.0 & \textbf{57.0} & 6.7 & 43.3 \\
\midrule
\rowcolor{blue!8}
\textbf{Multi-Task ES+SS (Ours)} & \textbf{70.6} & \textbf{98.0} & 44.5 & \textbf{14.0} & \textbf{56.8} \\
\bottomrule
\end{tabular}%
}
\end{wraptable}

\textbf{Fine-tuning choices.} \added{In Table \ref{tab:robotwin_ft_main}, we find full fine-tuning $\pihalf$ with Eq. \ref{eq:dual_objective} loss performs best as compared to LoRA \citep{hu2021lora} or selectively freezing VLM, SigLIP vision encoder, or action expert weights (Appendix~\ref{app:ft_ablations}).}

\begin{table}[t]
\begin{minipage}[t]{0.52\textwidth}
\centering\scriptsize
\renewcommand{\arraystretch}{0.95}\setlength{\tabcolsep}{3pt}
\caption{\textbf{Blind user study on held-out skills.} \emph{Multi-Task ES+ES (Oracle)} $\rightarrow$ \emph{Multi-Task ES+SS (Ours)}; 6 raters $\times$ 20 scenes per task (\%). All differences except IF have $p<0.001$ (exact McNemar; sign test for preference, ties omitted).}
\label{tab:user_study}
\resizebox{\linewidth}{!}{%
\begin{tabular}{@{}l cc@{}}
\toprule
\textbf{Metric} & \textbf{Push} & \textbf{Pour / turn over} \\
\midrule
IF (correct object)            & 82 $\rightarrow$ 84 & 91 $\rightarrow$ 97 \\
Attempted the requested motion & 38 $\rightarrow$ \textbf{82} & 43 $\rightarrow$ \textbf{94} \\
Picks up instead $\downarrow$  & 43 $\rightarrow$ \textbf{16} & --- \\
SR (correct object)            & 22 $\rightarrow$ \textbf{71} & 18 $\rightarrow$ \textbf{51} \\
Preferred by raters            & \phantom{0}9 $\rightarrow$ \textbf{68} & 11 $\rightarrow$ \textbf{75} \\
\bottomrule
\end{tabular}%
}

\vspace{2pt}
\caption{\textbf{Self-demos help even when expert data covers the task.} Adding pick-place self-demos to expert pick-place demos (``Pick up X and place it in Y''); blind A/B on the 20 $T_{SS}$ scenes. (\%)}
\label{tab:es_vs_esss_pnp}
\resizebox{\linewidth}{!}{%
\begin{tabular}{@{}l ccc@{}}
\toprule
\textbf{Method} & IF (pre-grasp) & IF (post-grasp) & SR \\
\midrule
Multi-Task ES (pick-place)               & 75 & \textbf{85} & 55 \\
\rowcolor{blue!8}
\textbf{Multi-Task ES+SS (pick-place)}   & \textbf{95} & 80 & \textbf{65} \\
\bottomrule
\end{tabular}%
}
\end{minipage}\hfill
\begin{minipage}[t]{0.44\textwidth}
\centering
\vspace{0pt}
\includegraphics[width=\linewidth]{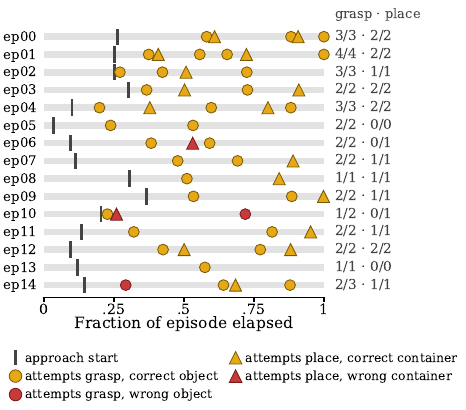}
\captionof{figure}{\textbf{Analysis of self-demos} used for training in Table~\ref{tab:real_robot_results}. 94\% of grasp and 89\% of place attempts target the prompted object, 13/15 episodes retry (mean 2.1 grasp and 1.2 place attempts per episode), but no grasp succeeds resulting in task failure.}
\label{fig:selfdemo_timeline}
\end{minipage}
\end{table}

\textbf{Scaling self-demos (sim).} We scaled up self-demos in simulation from 5 to 50 per task without filtering. Average performance does not change substantially as we scale up self-demos. But retention is better on tasks where base RoboTwin policy does well and regresses on tasks where the base policy does not do well. The prior's presence matters more than the quantity.

\textbf{Self-demo rollouts supply states, not just targets.} 
We conduct an ablation on the \emph{Multi-Task ES+ES} experiment where we query the frozen base $\pihalf$ on expert pick-place demo states, and fine-tune on these offline chunks instead of trajectory actions. We find 0\% SR. This could be due to the expert rollout states being out-of-distribution for the base policy and per-timestep action chunks being temporally inconsistent.

\textbf{Failure modes.} Our method can sometimes inherit artifacts from the data it rehearses. For example, since neither the teleoperation nor the generative prior data shows the scene \emph{after} a successful place, the policy is never taught when to stop, sometimes re-picks the placed object (Figure \ref{fig:montage_failure_mode_rollouts}).

\textbf{Limitations.} We find two experiments where the embodiment gap propagated to the fine-tuned policy: \added{(i) on the caterpillar checkpoint, \emph{Multi-Task ES+SS} retains instruction following, but it is worse at completing grasps for pick-place objects, possibly because the gears (from expert data) have larger radius than the standard pick-place objects (in self-demos) and it has been trained only on noisy grasp attempts for them (Table \ref{tab:caterpillar_pickplace}, Appendix \ref{app:caterpillar_pick_place}).} (ii) We conducted a control experiment on \emph{Multi-task ES (pick-place) + SS (pick-place)} from Table \ref{tab:es_vs_esss_pnp} with lower quality self-demos (poor semantic filtering) and the performance dropped as compared to the expert-only. Currently, selecting prompts $\mathcal{P}_{\text{SS}}$ for self-supervised data collection and filtering on the target robot requires manual curation, and can be automated in the future for larger-scale self-supervised learning. We evaluate only $\pihalf$, but our post-training recipe can easily be extended to other robotic foundation models.

\section{Conclusion and Future Work}

Pretrained VLAs suffer from embodiment gaps on a new robot even under minor differences in camera placement, gripper, etc, and re-collecting data for every pretrained skill on that robot is expensive. We show the base policy can supply priors from its own pretraining: rolled out on the target robot, its own generated actions serve as rehearsal targets that need no access to pretraining data and the resultant states carry no domain gap. 
Self-supervised imitation learning helps mitigate forgetting across both simulation and real world results. This is a particularly surprising result in the real world. The self-supervised data mostly has failures, and yet training on them recovers the self-supervised tasks, improves the expert tasks, and preserves held-out skills. VLA pretraining thus serves not only as initialization but as a prior generator. Future works can extend this into a data flywheel, distilling successive rounds of policy-generated demos mixed with expert data. 
We hope this is a step towards making pretrained VLAs practical to adapt without losing generalization, especially for the broader robotics community that cannot pretrain one from scratch.

\label{sec:conclusion}

\subsection*{AI use statement}
In this work, we used generative AI tools for finding relevant papers in parallel literatures, in code debugging, and generating visualizations and plots during project development.

\subsubsection*{Acknowledgments}
We thank the Siebel School Robotics Lab, UIUC for access to the stationary ALOHA-1 robot platform. We thank Amin Mirzaee for help 3D printing new gripper finger mounts and test set replacement objects, Trossen Robotics engineers for various hardware and system repair discussions, and Sanjay Pokkali for initial help with teleoperation. Thanks to Ansel Blume, Yuqun Wu, Aman Mehra, and Ruihan Gao for feedback on drafts. Prachi was supported by ONR award N00014-23-1-2383 and NSF award IIS 23-12102. This research project has benefited from the Microsoft Agentic AI Research and Innovation (AARI) grant program for Azure credits. We are also grateful for NVIDIA H200 GPUs through the Delta system at the National Center for Supercomputing Applications [award OAC 2005572] through allocation CIS240213 from the Advanced Cyberinfrastructure Coordination Ecosystem: Services \& Support (ACCESS) program, which is supported by National Science Foundation grants \#2138259, \#2138286, \#2138307, \#2137603, and \#2138296.

\bibliography{references}
\bibliographystyle{iclr2027_conference}

\clearpage
\appendix

\section*{Fine-Tuning VLAs with Self-Demonstrated Generative Control for Multi-Task Manipulation}

\section*{Appendix Overview}

This appendix provides benchmark and implementation details, additional ablation experiments, and qualitative results supplementing the main paper. It is organized as follows:

\begin{itemize}
    \item \textbf{Appendix~\ref{app:robotwin_exp}}: RoboTwin experiments
    \begin{itemize}
        \item Rehearsal mixing and buffer size ablations
        \item Selective fine-tuning ablations
        \item Frame-wise vs.\ episode-wise rehearsal upper bounds
        \item Knowledge insulation experiments for mid-training
        \item $T_{NO}$/$T_{NC}$ OOD benchmark results
    \end{itemize}
    \item \textbf{Appendix~\ref{app:caterpillar_pick_place}}: Caterpillar benchmark $T_{SS}$ ``pick-and-place" results
    \item \textbf{Appendix~\ref{app:user_study}}: User study details
    \item \textbf{Appendix~\ref{app:impl}}: Implementation details
    \begin{itemize}
        \item Infrastructure
        \item ALOHA robot: hyperparameters, data collection, and self-demo prompts
        \item RoboTwin: embodiments (E1, E2), customized tasks, hyperparameters, and seeds
    \end{itemize}
    \item \textbf{Appendix~\ref{app:benchmark_details}}: Benchmark details
    \begin{itemize}
        \item Benchmark details: train data, test set curation
        \item Success metric definitions
        \item RoboTwin task descriptions
        \item Real ALOHA benchmark test set descriptions and object configuration visualizations
    \end{itemize}
    \item \textbf{Appendix~\ref{app:real_rollouts}}: ALOHA robot qualitative results
    \begin{itemize}
        \item Rollout analysis
        \item Failure modes of self-supervised distillation (Figure~\ref{fig:montage_failure_mode_rollouts})
    \end{itemize}
\end{itemize}

\section{Additional RoboTwin Experiments}
\label{app:robotwin_exp}

\subsection{Rehearsal Mixing and Buffer Size}
\begin{table}[h]
\centering
\caption{\textbf{Rehearsal buffer ablations, RoboTwin validation set (E1).} SS:ES is the self-demo (rehearsal) to expert data ratio per batch on average: 1:1 samples equally when constructing batches, natural-ratio sampling draws in proportion to dataset sizes instead of fixing the ratio. Self-demos per task indicates dataset size the sampling is done from. SS sampling: Uniform draws every self-demo frame equally often based on natural episode lengths of various tasks, task-balanced weights frames inversely to their task's frame count so every $T_{SS}$ task is drawn equally often. Average SR (\%, $\uparrow$).}
\label{tab:robotwin_sampling_buffer_ablation}
\renewcommand{\arraystretch}{1.15}
\small
\begin{tabular}{@{}l ccc c c c@{}}
\toprule
\textbf{Ablation} & \shortstack{Self-demos\\per task} & \shortstack{SS:ES} & \shortstack{SS\\sampling} & $T_{SS}$ & $\Delta T_{SS}$ & $T_{ES}$ \\
\midrule
\rowcolor{gray!10}
\textbf{Multi-Task ES+SS (Ours)} & 10 & 1:1 & Uniform & \textbf{69.2} &  & 97.0 \\
\midrule
Task-balanced sampling & 10 & 1:1 & \cellcolor{orange!18}Task-balanced & 66.2 & $-$3.0 & 99.0 \\
Smaller self-demo buffer & \cellcolor{orange!18}5 & 1:1 & Uniform & 56.8 & $-$12.4 & 96.0 \\
\quad + natural-ratio sampling & \cellcolor{orange!18}5 & \cellcolor{orange!18}Proportional ($\sim$1:2) & Uniform & 56.4 & $-$12.8 & 97.0 \\
\bottomrule
\end{tabular}
\end{table}

Self-demos act as generative replay~\cite{shin2017continual}, the base policy supplies the data that preserves its own behavior. Across our sim and real experiments, three properties of this rehearsal matter. Self-demos must carry priors for the tasks being distilled from the base policy, though they may not necessarily succeed. There are several ablations that indicate this. On the robot experiments, self-demos are filtered for semantically relevant behavior and safety, and in sim they are unfiltered. \emph{Multi-Task ES+SS (Ours)} uses 10 self-demos per mid-train task and the default 50 expert demos for the post-train ones, giving the same episode budget in the rehearsal (self-demo) buffer and new (expert) tasks being learned. On average, each post-training mini-batch is constructed by sampling the two buffers 1:1. Halve the number of self-demo episodes (5 per task), SS:ES becomes 1:2. We conducted two ablations with this setting. \emph{Smaller self-demo buffer, Table \ref{tab:robotwin_sampling_buffer_ablation}} still constructs batches 1:1, oversampling from the smaller rehearsal buffer and suffers a 12.4 points drop in $T_{SS}$ as compared to Ours. Even using \emph{a smaller self-demo buffer + natural-ratio sampling} proportional to dataset sizes, $\sim1:2$ for constructing batches yields a drop of 12.8 points. This echoes overfitting to a small replay memory in rehearsal-based continual learning literature~\cite{chaudhry2019tiny}. 

\begin{figure}[!b]
\centering
\includegraphics[width=0.98\linewidth]{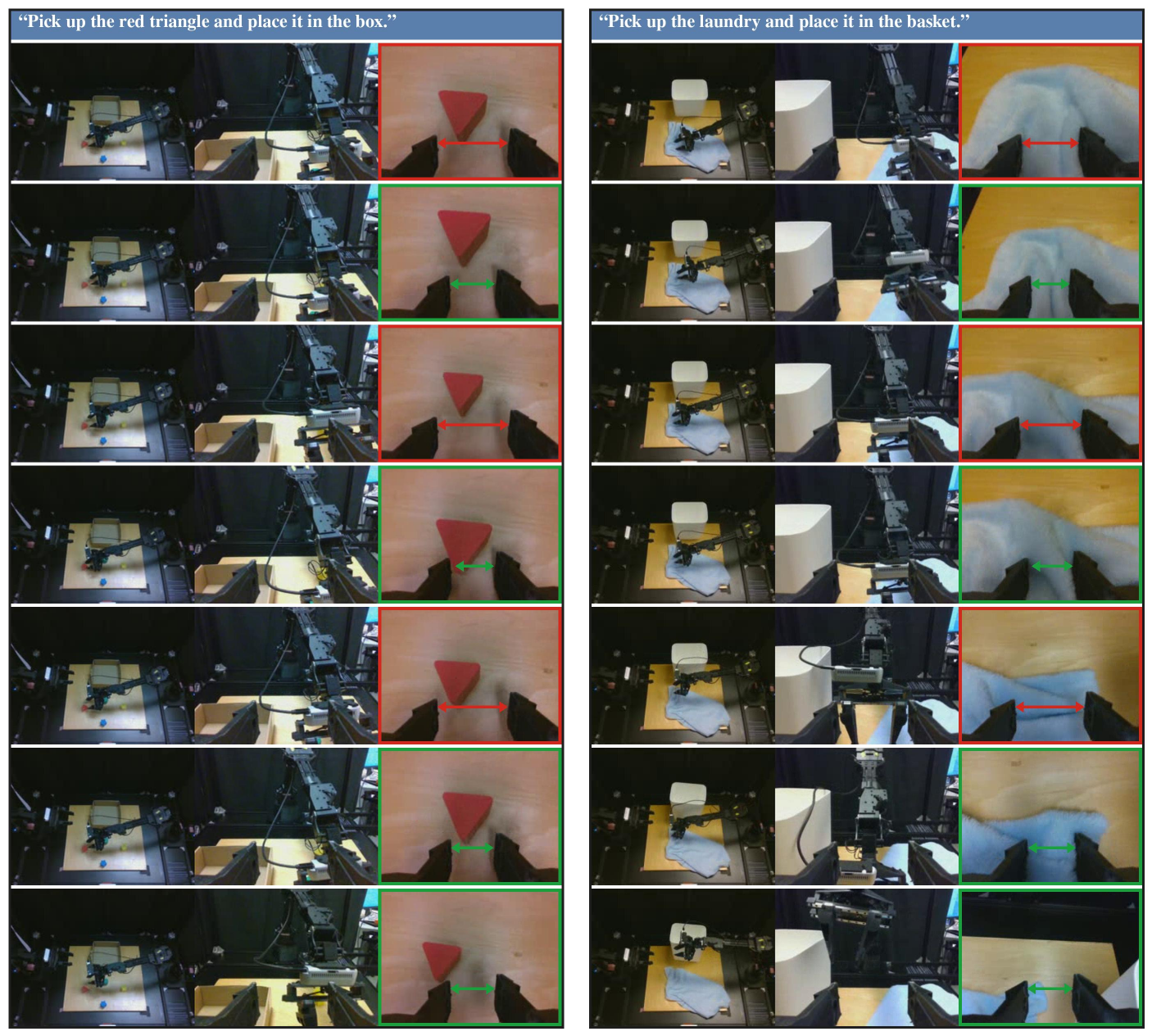}
\caption{\textbf{Rehearsal buffer ablation: too small a self-demo buffer.} \added{ Real-robot rollout of a model trained with ${\sim}$0.33$\times$
as many self-demo frames as expert frames, mixed 1:1. Columns: top, left-wrist
and right-wrist cameras. The right-wrist border marks the gripper closed (green) or reopened (red): the policy
repeatedly closes on the object and lets go instead of lifting it.}}
\label{fig:gripper_oscillation}
\end{figure}

On the real robot, we find that using smaller self-demo buffer data sizes or smaller/larger ones without 1:1 batch sampling leads to erratic grasping behavior for the $T_{SS}$ pick-and-place tasks such as constantly opening and closing the gripper which is not observed with \emph{Ours}. Figure \ref{fig:gripper_oscillation} shows one such example for rehearsal buffer one third of the expert sampled 1:1. Hence, the best results are attained with 1:1 sampling and having self-demo dataset sizes greater than or equal to expert data. The 1:1 sampling is in line with standard experience replay~\cite{chaudhry2019tiny,rolnick2019experience} and can be seen as distilling the priors with expert data from target embodiment in each batch. 

Additionally, \emph{Task-balanced sampling} from self-demo buffer instead of uniformly sampling based on the natural episode length across diverse tasks drops on $T_{SS}$ by 3 points when drawing each $T_{SS}$ task with probability inversely proportional to its frame count, up-weighting shorter trajectory tasks. 

\subsection{Fine-tuning ablations}
\label{app:ft_ablations}
\begin{table}[h]
\centering
\caption{\textbf{RoboTwin fine-tuning ablations (E1).} This table compares how \emph{Multi-Task ES} is post-trained from the RoboTwin Base Policy, extension of Table \ref{tab:robotwin_ft_main}. Parameter-efficient variants retain somewhat more $T_{SS}$ but learn the new tasks ($T_{ES}$) or generalize ($T_{NO}$) less well. \emph{Ours} adds self-demos to the selected setup (full fine-tuning with the FAST+Flow objective). Average success rates (\%, $\uparrow$); Overall is the mean over the four test sets.}
\label{tab:robotwin_ablations}
\renewcommand{\arraystretch}{1.1}
\small
\begin{tabular}{@{}c@{\hspace{4pt}}l ccccc@{}}
\toprule
& \textbf{Experiment} & $T_{SS}$ (old) & $T_{ES}$ (new) & $T_{NO}$ & $T_{NC}$ & \textbf{Overall} \\
\midrule
 & LoRA & 28.0 & 87.0 & 37.5 & 2.0 & 38.6 \\
 & Freeze SigLIP+VLM & 28.8 & 38.0 & 15.5 & 4.0 & 21.6 \\
 & Freeze VLM & 18.8 & 38.0 & 22.0 & 1.3 & 20.0 \\
\multirow{-4}{*}{\rotatebox{90}{\scriptsize\itshape PEFT (ES)}} & Freeze SigLIP & 23.8 & 93.0 & 46.5 & 5.3 & 42.2 \\
\cmidrule(l){1-7}\addlinespace[3pt]
 & Multi-Task ES (Flow-only) & 11.2 & 94.0 & \textbf{57.0} & 5.3 & 41.9 \\
\rowcolor{gray!10}
\multirow{-2}{*}{\rotatebox{90}{\scriptsize\itshape \shortstack{Full FT\\(ES)}}} & Multi-Task ES & 16.6 & 93.0 & \textbf{57.0} & 6.7 & 43.3 \\
\midrule
\rowcolor{blue!8}
\cellcolor{white} & \textbf{Multi-Task ES+SS (Ours)} & \textbf{70.6} & \textbf{98.0} & 44.5 & \textbf{14.0} & \textbf{56.8} \\
\bottomrule
\end{tabular}
\end{table}

As mentioned in section \ref{expert-supervision method}, ablations on selective freezing and LoRA fine-tuning confirm \emph{full fine-tuning performs best}. 
Unlike $\pi_0$'s flow-matching loss alone \citep{black2024pi_0}, $\pihalf$ adds a FAST autoregressive loss (Eq.~\ref{eq:dual_objective}) for knowledge insulation \citep{intelligence2025pi_05, driess2026knowledge}, omitted in OpenPI. Comparing against \emph{Multi-Task ES (Flow-only)} confirms FAST loss reduces $T_{SS}$ forgetting, matching our real-robot observations. We therefore use full fine-tuning with the dual objective for all real-robot experiments. 
While full fine-tuning performs best overall, it is recommended to tune at least VLM and action expert for learning new tasks, Table \ref{tab:robotwin_ablations}. 

\providecommand{\rhv}[1]{\rotatebox[origin=lB]{90}{#1}}
\begin{table*}[p]
\centering
\caption{\textbf{Full RoboTwin simulation results, self-supervised tasks $T_{SS}$ and expert tasks $T_{ES}$ test sets.} Mid-training rows are base policies trained on the 10 $T_{SS}$ tasks; all post-training rows are initialized from the $\pi_{0.5}$ RoboTwin Base Policy. Oracle rows (gray) assume access to stored mid-training data and are upper-bound references. Bold: best result without stored mid-training data. Success rates (\%, $\uparrow$) averaged across 50 seeds per task. Full version of Table~\ref{tab:robotwin_main_ES_result}.}
\label{tab:robotwin_ES_SS_full}\label{tab:stage1_test}
\renewcommand{\arraystretch}{1.1}
\setlength{\tabcolsep}{2.6pt}
\footnotesize
\resizebox{\textwidth}{!}{%
\begin{tabular}{@{}c@{\hspace{4pt}}l *{10}{>{\centering\arraybackslash}p{0.7cm}} | >{\centering\arraybackslash}p{0.7cm} | >{\centering\arraybackslash}p{0.7cm} >{\centering\arraybackslash}p{0.7cm} | >{\centering\arraybackslash}p{0.7cm}@{}}
\toprule
& & \multicolumn{11}{c|}{\textbf{$T_{SS}$: Mid-training tasks (old)}} & \multicolumn{3}{c}{\textbf{\shortstack{$T_{ES}$: Post-training\\tasks (new)}}} \\
\cmidrule(lr){3-13}\cmidrule(l){14-16}
& \textbf{Experiment} & \rhv{Place Dual Shoes} & \rhv{Beat Block Hammer} & \rhv{Pick Dual Bottles} & \rhv{Place Container Plate} & \rhv{Handover Mic} & \rhv{Place Empty Cup} & \rhv{Place Bread Basket} & \rhv{Lift Pot} & \rhv{Open Microwave} & \rhv{Put Bottles Dustbin} & \rhv{\textbf{Avg. $T_{SS}$}} & \rhv{Stack $R \rightarrow G$} & \rhv{Stack $G \rightarrow R$} & \rhv{\textbf{Avg. $T_{ES}$}} \\
\midrule
 & $\pi_{0.5}$ Zero-Shot & -- & -- & -- & -- & -- & -- & -- & -- & -- & -- & 0.0 & -- & -- & -- \\
\multicolumn{16}{@{}l}{\makebox[0pt][l]{\textbf{\textit{Mid-Training Experiments}}}} \\[2pt]
 & Flow-only (openpi default) & 72 & 100 & 100 & 100 & 100 & 98 & 82 & 98 & 52 & 94 & 89.6 & -- & -- & -- \\
 & FAST+Flow+KI & 2 & 34 & 42 & 90 & 30 & 78 & 28 & 66 & 12 & 25 & 40.7 & -- & -- & -- \\
 & $\pi_{0.5}$ RoboTwin Base Policy & 76 & 94 & 98 & 100 & 100 & 98 & 82 & 98 & 72 & 90 & 90.8 & -- & -- & -- \\
\midrule
\multicolumn{16}{@{}l}{\makebox[0pt][l]{\textbf{\textit{Post-Training Experiments: All models initialized from RoboTwin Base Policy}}}} \\[2pt]
\rowcolor{gray!12}
\cellcolor{white} & Rehearsal & 58 & 92 & 100 & 96 & 100 & 94 & 86 & 92 & 56 & 82 & 85.6 & 96 & 98 & 97.0 \\
\rowcolor{gray!12}
\cellcolor{white} & Rehearsal + LoRA & 74 & 94 & 96 & 96 & 94 & 94 & 90 & 92 & 40 & 76 & 84.6 & 82 & 72 & 77.0 \\
\rowcolor{gray!12}
\cellcolor{white}\multirow{-3}{*}{\rotatebox{90}{\scriptsize\itshape Oracles}} & Rehearsal, episode-wise & 64 & 88 & 90 & 98 & 88 & 90 & 76 & 100 & 50 & 74 & 81.8 & 100 & 100 & 100.0 \\
\cmidrule(l){1-16}
 & LoRA & 4 & 4 & 18 & 76 & 0 & 46 & 50 & 76 & 4 & 2 & 28.0 & 84 & 90 & 87.0 \\
 & Freeze SigLIP+VLM & 2 & 0 & 28 & 78 & 58 & 26 & 34 & 40 & 16 & 6 & 28.8 & 32 & 44 & 38.0 \\
 & Freeze VLM & 2 & 0 & 4 & 70 & 0 & 26 & 34 & 36 & 14 & 2 & 18.8 & 40 & 36 & 38.0 \\
\multirow{-4}{*}{\rotatebox{90}{\scriptsize\itshape PEFT (ES)}} & Freeze SigLIP & 0 & 2 & 4 & 60 & 0 & 76 & 30 & 62 & 4 & 0 & 23.8 & 94 & 92 & 93.0 \\
\cmidrule(l){1-16}\addlinespace[3pt]
 & Multi-Task ES (Flow-only) & 0 & 0 & 0 & 46 & 0 & 24 & 16 & 0 & 26 & 0 & 11.2 & 94 & 94 & 94.0 \\
\multirow{-2}{*}{\rotatebox{90}{\scriptsize\itshape \shortstack{Full FT\\(ES)}}} & Multi-Task ES$^\dagger$ & 0 & 0 & 0 & 60 & 0 & 36 & 34 & 36 & 0 & 0 & 16.6 & 90 & 96 & 93.0 \\
\midrule
\rowcolor{blue!8}
\cellcolor{white} & \textbf{Multi-Task ES+SS (Ours)} & \textbf{42} & \textbf{82} & \textbf{48} & \textbf{92} & \textbf{62} & \textbf{92} & \textbf{58} & \textbf{86} & \textbf{66} & \textbf{78} & \textbf{70.6} & \textbf{98} & \textbf{98} & \textbf{98.0} \\
\bottomrule
\end{tabular}%
}
\vspace{1.5em}

\centering
\caption{\textbf{RoboTwin novel objects $T_{NO}$ and novel task composition $T_{NC}$ test sets.} Generalization of the $T_{ES}$ stack two blocks task to (i) new objects cylinder and sphere and (ii) composing new tasks, stacking bowls and stacking three cubes. Bold: best result without stored mid-training data. Success rates (\%, $\uparrow$) averaged across 50 seeds per task.}
\label{tab:robotwin_stage2_ood_results}
\renewcommand{\arraystretch}{1.1}
\setlength{\tabcolsep}{3.5pt}
\footnotesize
\resizebox{0.76\linewidth}{!}{%
\begin{tabular}{@{}c@{\hspace{4pt}}l *{4}{c} | c | >{\centering\arraybackslash}p{0.62cm} >{\centering\arraybackslash}p{0.62cm} >{\centering\arraybackslash}p{0.62cm} | >{\centering\arraybackslash}p{0.62cm}@{}}
\toprule
& & \multicolumn{5}{c|}{\textbf{$T_{NO}$: Novel Object Tasks}} & \multicolumn{4}{c}{\textbf{\shortstack{$T_{NC}$: Novel\\Composition Tasks}}} \\
\cmidrule(lr){3-7}\cmidrule(l){8-11}
& \textbf{Experiment} & \rhv{Cylinder on Cube} & \rhv{Sphere on Cube} & \rhv{Sphere + Distractor} & \rhv{Cylinder + Distractor} & \rhv{\textbf{Avg. OOD objects}} & \rhv{Stack Blocks Three} & \rhv{Stack Bowls Two} & \rhv{Stack Bowls Three} & \rhv{\textbf{Avg. OOD compose}} \\
\midrule
\multicolumn{11}{@{}l}{\makebox[0pt][l]{\textbf{\textit{Post-Training Experiments: All models initialized from RoboTwin Base Policy}}}} \\[2pt]
\rowcolor{gray!12}
\cellcolor{white} & Rehearsal & 42 & 46 & 12 & 20 & 30.0 & 2 & 44 & 0 & 15.3 \\
\rowcolor{gray!12}
\cellcolor{white} & Rehearsal + LoRA & 44 & 40 & 14 & 20 & 29.5 & 0 & 34 & 0 & 11.3 \\
\rowcolor{gray!12}
\cellcolor{white}\multirow{-3}{*}{\rotatebox{90}{\scriptsize\itshape Oracle}} & Rehearsal, episode-wise & 62 & 68 & 18 & 30 & 44.5 & 0 & 48 & 2 & 16.7 \\
\cmidrule(l){1-11}
 & LoRA & 46 & 66 & 22 & 16 & 37.5 & 0 & 6 & 0 & 2.0 \\
 & Freeze SigLIP+VLM & 22 & 26 & 10 & 4 & 15.5 & 0 & 12 & 0 & 4.0 \\
 & Freeze VLM & 32 & 32 & 10 & 14 & 22.0 & 0 & 4 & 0 & 1.3 \\
\multirow{-4}{*}{\rotatebox{90}{\scriptsize\itshape PEFT (ES)}} & Freeze SigLIP & 58 & \textbf{74} & 30 & 24 & 46.5 & 0 & 16 & 0 & 5.3 \\
\cmidrule(l){1-11}\addlinespace[3pt]
 & Multi-Task ES (Flow-only) & \textbf{78} & 70 & 38 & \textbf{42} & \textbf{57.0} & 0 & 14 & \textbf{2} & 5.3 \\
\multirow{-2}{*}{\rotatebox{90}{\scriptsize\itshape \shortstack{Full FT\\(ES)}}} & Multi-Task ES$^\dagger$ & 76 & 64 & \textbf{46} & \textbf{42} & \textbf{57.0} & 0 & 20 & 0 & 6.7 \\
\midrule
\rowcolor{blue!8}
\cellcolor{white} & \textbf{Multi-Task ES+SS (Ours)} & 64 & 56 & 26 & 32 & 44.5 & 0 & \textbf{42} & 0 & \textbf{14.0} \\
\bottomrule
\end{tabular}
}
\end{table*}

\subsection{Frame-wise vs.\ episode-wise rehearsal}
We compare three upper-bound baselines that store expert data from mid-training and rehearse it during post-training instead of using self-demos. Prior work using experience replay for continual robotic manipulation stores a fixed budget of randomly sampled frames per task \citep{liu2023libero, liu2026vla}. Our \emph{Rehearsal} and \emph{Rehearsal + LoRA} \citep{liu2026vla} baselines follow this frame-wise setup, storing 20\% of mid-training frames. We additionally store whole episodes instead, using the first 10 episodes per task (20\% of episodes). \emph{Rehearsal, episode-wise} achieves higher $T_{ES}$, $T_{NO}$, and $T_{NC}$ (100.0, 44.5, 16.7 vs.\ 97.0, 30.0, 15.3) at slightly lower $T_{SS}$ (81.8 vs.\ 85.6), giving a higher Overall (60.8 vs.\ 57.0) (see Tables \ref{tab:robotwin_ES_SS_full}, \ref{tab:robotwin_stage2_ood_results}). We leave a systematic comparison of rehearsal buffer curation strategies to future work.  

\subsection{Mid-training with knowledge insulation}
The $\pihalf$ model checkpoint released publicly as part of the openpi code is pretrained with the FAST-Flow dual objective Eq. \ref{eq:dual_objective} and knowledge insulation (KI) \citep{driess2026knowledge}, but the code release excludes FAST and KI implementation. In Table \ref{tab:robotwin_ES_SS_full}, we experimented \emph{FAST+Flow+knowledge insulation (KI)} fine-tuning for RoboTwin mid-training stage, such that a stop gradient prevents flow-gradients from updating VLM's weights. The FAST loss mid-trains with only the VLM backbone and Flow loss tunes the action expert. This was found to perform worst, indicating KI is perhaps more suitable for VLM to VLA fine-tuning when the action expert is initialized from scratch. The mid-training conducted with FAST+Flow dual objective is the \emph{$\pihalf$ RoboTwin Base Policy}.

\subsection{RoboTwin \textbf{$T_{NO}$, $T_{NC}$} test sets benchmark}
Table \ref{tab:robotwin_ES_SS_full} shows all 10 tasks from Table \ref{tab:robotwin_main_ES_result}. In Table \ref{tab:robotwin_stage2_ood_results}, we show that our method (\emph{Multi-Task ES+SS}) outperforms all baselines and is only behind the frame-wise oracle by 1.33\% for three tasks in $T_{NC}$. It is the best performing method that does not assume access to or rely on any data from mid-training for retaining generalization. The \emph{Multi-Task ES} baseline is the best for object generalization in $T_{NO}$. Note, the nature of the base policy's pretraining tasks is different across simulation and real benchmarks, in real, our self-demos are for picking and placing different objects which also helps improve performance on $T_{NO}$. Whereas in simulation, our pretraining tasks are very different from the tasks in $T_{NO}$, $T_{NC}$ (open microwave, handover mic, pick-place different objects). We don't optimize for either of these test sets and release this for completeness of our generalization benchmark, consistency with real experiments, and future use.    

\section{Caterpillar gear insertion extended results}
\label{app:caterpillar_pick_place}

\begin{table}[h]
\centering
\caption{\textbf{ALOHA Robot caterpillar.} Policies fine-tuned on gear insertion, evaluated on $T_{SS}$ pick-and-place test set. \emph{Multi-Task ES} and \emph{Multi-Task ES+SS} were scored blind on the same 20 scenes; the zero-shot row is from Table~\ref{tab:real_robot_results}. \emph{Held in target}: the rollout ends with the object held by the gripper inside the container, hence not counted in SR. 
McNemar $p$-value on discordant pairs.}
\label{tab:caterpillar_pickplace}
\renewcommand{\arraystretch}{1.15}
\resizebox{0.9\textwidth}{!}{%
\begin{tabular}{@{}l @{\hspace{1.0em}} *{5}{c}@{}}
\toprule
& \multicolumn{5}{c}{\textbf{$T_{SS}$ Self-Supervised Tasks}} \\
& \multicolumn{5}{c}{\textit{(``Pick up X and place it in Y'')}} \\
\cmidrule(lr){2-6}
\textbf{Method}
& IF (pre-grasp) & Pick SR & IF (post-grasp) & Held in target & SR \\
\midrule
$\pihalf$ Zero-Shot                        & 70 & -- & 25 & -- & 0 \\
\midrule
$\pihalf$ Multi-Task ES                    & \textbf{90} & \textbf{85} & \textbf{85} & 25 & 0 \\
\rowcolor{blue!8}
\textbf{$\pihalf$ Multi-Task ES+SS (Ours)} & 80 & 20 & 35 & 15 & 0 \\
\midrule
\textit{$p$ (Ours vs.\ ES)}          & $>$0.1 & $<$0.001 & 0.002 & $>$0.1 & $>$0.1 \\
\bottomrule
\end{tabular}%
}
\end{table}

\label{app:caterpillar_ab}
\added{We compare the caterpillar \emph{Multi-Task ES+SS} vs \emph{Multi-Task ES} checkpoints blind on $T_{SS}$ pick-and-place test set (Table~\ref{tab:caterpillar_pickplace}). 
Clips were anonymized and shuffled, and scored before identities were revealed. Pre-grasp instruction following is similar (90\% vs.\ 80\%), but \emph{Multi-Task ES} picks more often (20\% vs.\ 85\%, $p<0.001$), and goes to place more often (post-grasp IF is 35\% vs.\ 85\%, $p=0.002$). Neither policy completes a place (0\% SR). Both sometimes hold the object inside the target container without releasing it.}

\section{User Study}
\label{app:user_study}

\begin{figure}[!htbp]
\centering
\includegraphics[width=0.49\linewidth]{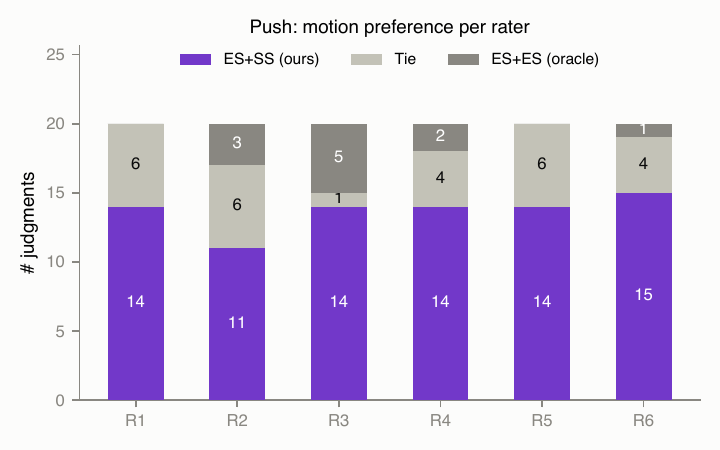}\hfill
\includegraphics[width=0.49\linewidth]{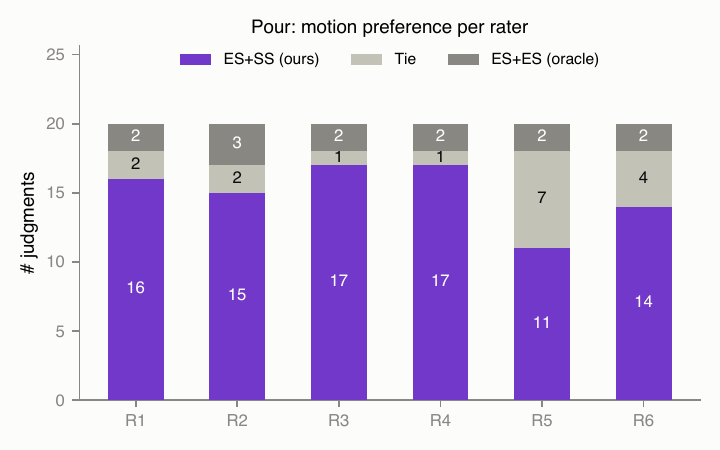}
\caption{\textbf{Blind preference by rater.} Each rater's 20 paired judgments per task: all six raters independently prefer ours over the all-expert oracle.}
\label{fig:user_study_by_rater}
\end{figure}

\begin{figure}[p]
\centering
\includegraphics[width=\linewidth,height=0.53\textheight,keepaspectratio]{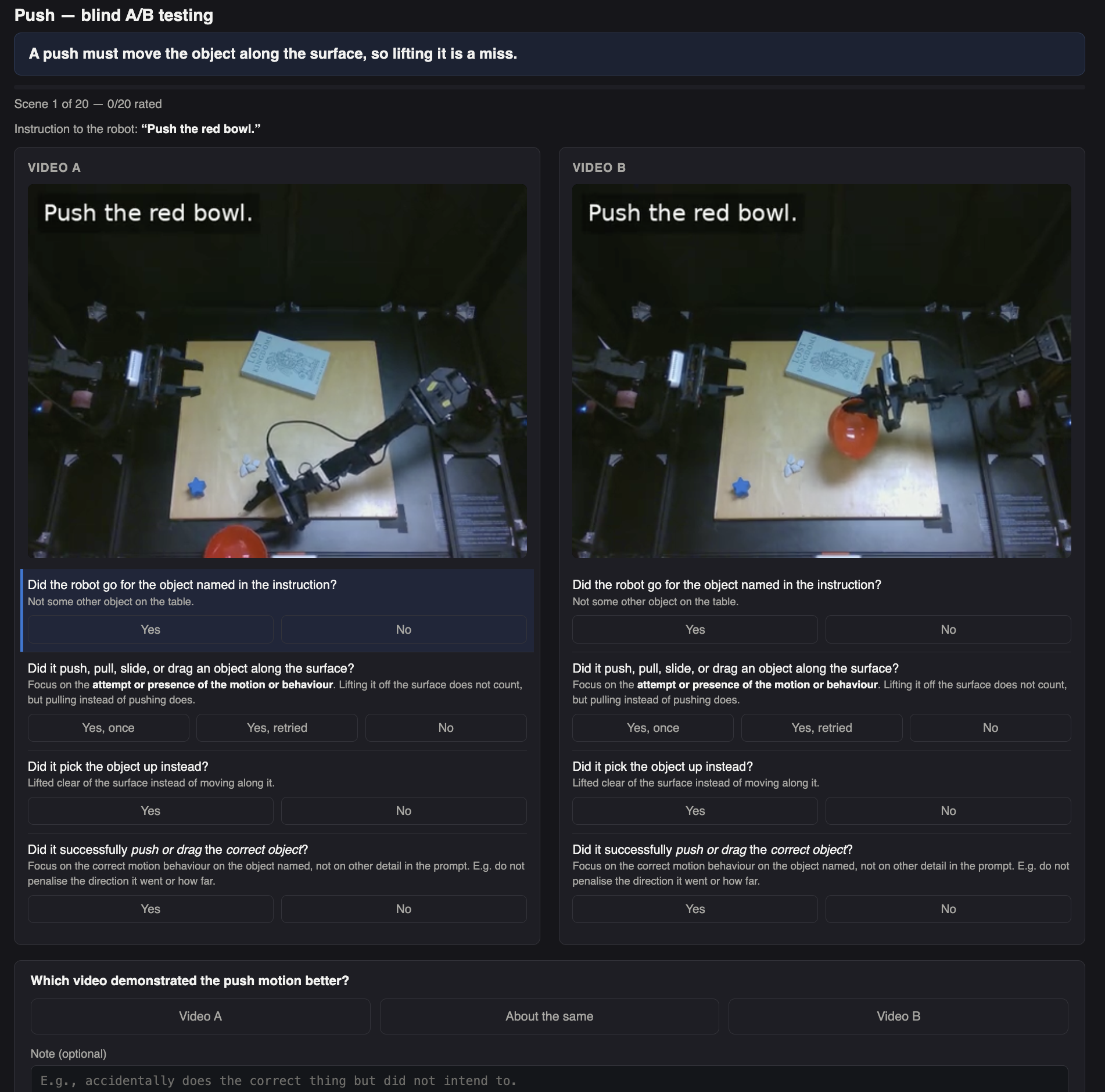}
\caption{The user study portal for push task.}
\label{fig:user_study_portal}
\vspace{0.4em}
\includegraphics[width=\linewidth,height=0.38\textheight,keepaspectratio]{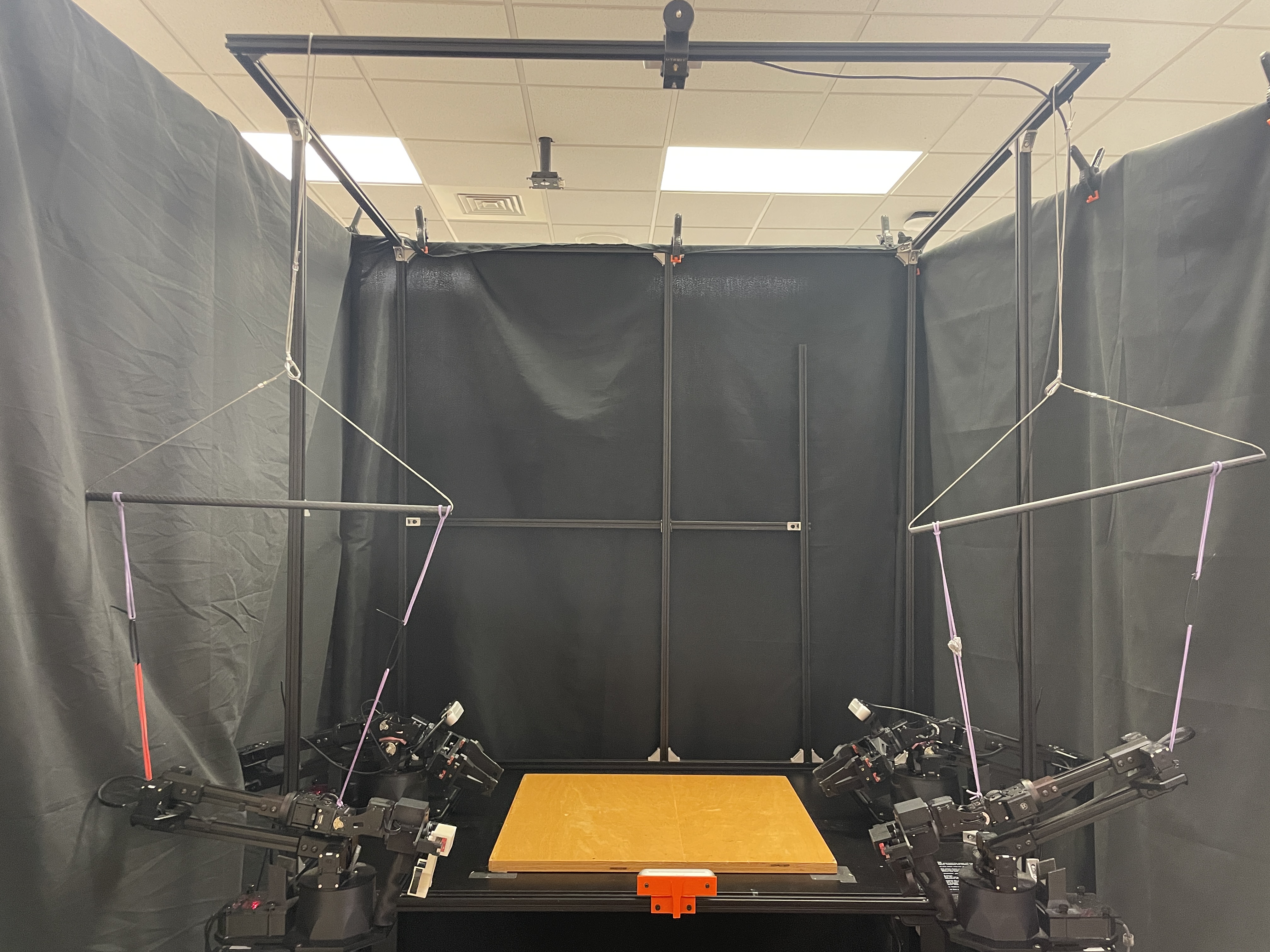}
\caption{Our stationary ALOHA-1 \citep{zhao2023learning} platform with two ViperX 300 follower arms and two WidowX 250 leader arms for collecting expert human demonstrations via teleoperation. It has parallel jaw grippers. We use RGB images from the top scene camera, left wrist camera, and the right wrist camera as input. All of these are Intel RealSense D435 cameras, with raw image resolution of $640 \times 480$ pixels.}
\label{fig:our-aloha-system}
\end{figure}

\added{To compare the oracle \emph{Multi-Task ES+ES} and our policy \emph{Multi-Task ES+SS} in Table \ref{tab:user_study} on held-out tasks without experimenter bias, 6 raters each judged 20 scenes per task (push, pour/turn-over). For every scene, both policies' rollouts were shown in randomized, policy-blind left/right order (interface in Figure~\ref{fig:user_study_portal}), and raters answered per-clip questions plus one paired preference question. The main results in Table \ref{tab:user_study} are carried by broad agreement across raters, not by a few outliers: Figure~\ref{fig:user_study_by_rater} shows that every rater independently preferred ours in a majority of scenes. Raters showed no left/right position bias (sign test $p=0.84$ push, $p=1.0$ pour), and inter-rater agreement on the preference question was moderate (Fleiss' $\kappa=0.45$ push, $0.37$ pour), with ours strongly preferred.}

\clearpage
\section{Implementation Details}
\label{app:impl}

\subsection{Infrastructure}
$\pihalf$ full fine-tuning requires 120+ GB GPU VRAM. All our simulation and real models were trained on clusters using $1 \times H200s$ having 140 GB VRAM each, or $2 \times H100s$ having 96 GB VRAM each. All RoboTwin evaluations were conducted on nodes with L40 GPUs. Our real ALOHA robot is attached to a NVIDIA GeForce RTX 4090, and we run both the client and policy server during inference on the same machine (though hosting the server remotely also works). All other development was conducted on another 4090 Desktop. Our hardware platform is described in Figure \ref{fig:our-aloha-system}. We collect raw data in HDF5 format using the original ALOHA Interbotix ROS-python code and convert it into HuggingFace LeRobot Parquet data format for the dataloader.  

\subsection{ALOHA Robot}

\textbf{Maximum steps used during inference.} For all test rollouts in `Pick up' family, we use max steps 800, 1000 for `Pick-and-place', 1200 for `Pick up laundry and place it in the basket', and 1650 for all caterpillar tasks. 

\textbf{Hyperparameters.} We use the FAST+Flow joint objective from Equation \ref{eq:dual_objective} for all real experiments (baselines and our method). All fine-tuning experiments in Table \ref{tab:real_robot_results} were trained for 10,000 iterations with a batch size of 64, with $\alpha=8.0$. The caterpillar dataset is much larger than the pick up tasks so we train all models for 15,000-17,000 iterations at a batch size of 64 and $\alpha=10.0$. 

\textbf{Base model definition.} We use the $\pihalf$ checkpoint released as part of openpi codebase as the zero-shot policy for all real experiments and for generating self-demos (which are all unsuccessful). 

\textbf{Benchmark expert teleoperation data collection.} For the \emph{Multi-Task ES} experiment, we collect data such that there are 8 unique scenes with red and green cube. For each of `Pick up the red cube' and `Pick up the green cube', we collect 2 demonstrations in each of 7 scenes and 1 in an eighth scene, giving 15 demonstrations per instruction and 30 in total. Both instructions use the identical set of scenes. We use this dataset for expert-supervision in \emph{Multi-Task ES} experiment. For the $\pihalf$ \emph{Single-Task ES} experiment in Table \ref{tab:real_robot_results}, we collect data from 15 unique scenes with both red and green cubes in them, but collecting data only for the `Pick up the green cube' task, with two demos from each scene for a dataset of 30 demonstrations. For the \emph{Multi-Task ES, Single Object/Scene} experiment, we collect 15 demos from `Pick up the red cube' and 15 from `Pick up the green cube', with only the corresponding cube in the scene. For a fair comparison, cube positions match those in the \emph{Multi-Task ES} dataset exactly; the only difference is that the distractor cube is absent.

For the caterpillar benchmark in Table \ref{tab:caterpillar}, we collected 60 demonstrations
with 20 demos each of orange, purple and green gears such that there are 2 or 3 unique instructions from each scene. The board is not fixed. The arm closest to the gear is used to pick up the gear and place it on the gear shaft while the other arm is used to help hold the caterpillar board to help align the gear if required. 

\textbf{Base policy self-demo data.} For our method in Table \ref{tab:real_robot_results}, we collect 15 self-demonstrations. 
Details of the prompts we used are in Table \ref{tab:self_demo_prompts}. For our method in Table \ref{tab:caterpillar}, we augment Table \ref{tab:self_demo_prompts} with more, using 29 self-demonstrations as this expert dataset is larger than the pick cubes expert data.

\begin{table}[!htbp]
\centering
\caption{Task prompts $\mathcal{P}_{\text{SS}}$ for collecting rehearsal data $\mathcal{D}_{\text{SS}}$.}
\label{tab:self_demo_prompts}
\small
\setlength{\tabcolsep}{2pt}
\resizebox{\textwidth}{!}{%
\begin{tabular}{@{}cl@{\hspace{1em}}cl@{}}
\toprule
\textbf{Ep.} & \textbf{Task Prompt} & \textbf{Ep.} & \textbf{Task Prompt} \\
\midrule
0 & Pick up the laundry and place it in the basket.             & 8  & Pick up the spoon and place it in the red bowl. \\
1 & Pick up the blue cloth and place it in the basket.          & 9  & Pick up the marker and place it in the blue bowl. \\
2 & Pick up the laundry and place it in the basket.             & 10 & Pick up the yellow object and place it in the mug. \\
3 & Pick up the laundry and place it in the basket.             & 11 & Pick up the blue star object and place it into the box. \\
4 & Pick up the laundry and place it in the basket.             & 12 & Pick up the spoon and place it in the blue bowl. \\
5 & Pick up the red triangular object and place it in the box. & 13 & Pick up the marker and put it into the gray box. \\
6 & Pick up the purple cube and place it in the red bowl.       & 14 & Go to the red cube, pick it up, and drop it in the brown box. \\
7 & Pick up the red triangular object and place it in the box. &    & \\
\bottomrule
\end{tabular}%
}
\end{table}

\newpage
\subsection{RoboTwin}
\label{sec:appendix-B-robotwin-customization}

\begin{figure}[h]
\centering
\includegraphics[width=\linewidth]{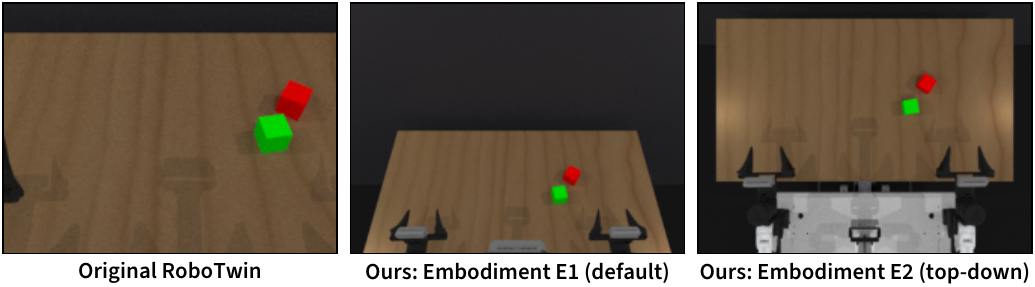}
\caption{\added{This shows the head camera's viewpoint in the original RoboTwin 2.0, ALOHA-Agilex configuration, and the two customized settings in which we run our experiments (Section \ref{sec:benchmark}).}}
\label{fig:robotwin-embodiments}
\end{figure}

\begin{wraptable}{r}{0.46\linewidth}
\vspace{-1em}
\centering
\caption{\added{Head-camera settings.}}
\footnotesize
\setlength{\tabcolsep}{4pt}
\begin{tabular}{lccc}
\toprule
 & RoboTwin & E1 & E2 \\
\midrule
$x$ (m)    & $-0.03$          & $0$              & $0$ \\
$y$ (m)    & $-0.45$          & $-0.55$          & $-0.10$ \\
$z$ (m)    & $1.35$           & $1.90$           & $1.90$ \\
Pitch      & $53.1^\circ$     & $53.1^\circ$     & $90^\circ$ \\
FoV        & $37^\circ$       & $43.8^\circ$     & $43.8^\circ$ \\
Resolution & $320{\times}240$ & $640{\times}480$ & $640{\times}480$ \\
\bottomrule
\end{tabular}
\label{tab:head_cameras}
\vspace{-1em}
\end{wraptable}

\textbf{ALOHA-Agilex Embodiment.} We customize RoboTwin's ALOHA-AgileX simulated embodiment to match our policy's I/O Figure \ref{fig:robotwin-embodiments}, Table \ref{tab:head_cameras}. The head and wrist cameras are switched to render at $640\times480$ (Large\_D435 profile), so simulated observations match $\pihalf$'s native $640\times480$ input resolution, with a wider field of view (fovy 43.8°). We vertically raise the head camera for an unobstructed task view with both arms, similar to our real setup. We use the cuRobo planner for collision-aware dual-arm trajectory generation and collect expert motion planner training data for mid-training and post-training tasks using this setup. $\pihalf$ randomly crops and resizes all camera images to a resolution of $224 \times 224$.

\textbf{Embodiment gap (E1 versus E2).} \added{E2 changes only the head camera, which is moved 0.45\,m forward over the table and points straight down, Table \ref{tab:head_cameras}. 
The robot, wrist cameras, camera model, scenes, tasks and test seeds are the same as in E1. }

\textbf{Customized tasks.} \added{Our $T_{ES}$ tasks, ``stack blocks red on green" and `stack blocks green on red" are two fixed-order variants of RoboTwin's `stack\_blocks\_two', Figure \ref{fig:montage_robotwin_tes}. $T_{NO}$ (Figure~\ref{fig:montage_robotwin_tno}) adds four variants of this task. Two replace one of the 5 cm cubes with a novel shape: a cylinder (radius 2.5 cm, height 5 cm) or a sphere (radius 2.5 cm). The other two also add a distractor cube of the same color as the novel object (cylinder/sphere), so the target must be identified by shape rather than color. Object colors are sampled per seed from 14 colors and fixed by the seed, released as part of our RoboTwin code. The prompt follows the $T_{ES}$ template (``Pick up the {c1} block and place it in the center of the table. Pick up the {c2} sphere/cylinder and stack it on top of the {c1} block."). Object placement, the scripted expert and the success check (top object within 2.5 cm in x/y and 1.2 cm in z of its stacked pose) are identical to `stack\_blocks\_two'. $T_{NC}$ uses RoboTwin's stock `stack\_blocks\_three', `stack\_bowls\_two' and `stack\_bowls\_three' without customization.}

\textbf{Hyperparameters.} We use a validation set to select how many iterations to train for and batch size for both mid-training and post-training. We find mid-training performance to be best at 10,000 iterations with batch size 128. We use this for reporting performance on test set. For post-training, we find that performance first drops for all tasks and then starts to improve, and select 6,000 iterations as the sweet spot between stability and plasticity on old and new task performance. All our RoboTwin post-training results are trained with batch size 64 for 6000 iterations with a post-warmup LR=2.5e-5. We found RoboTwin mid-training to be very sensitive to class imbalance and the FAST loss does not converge if there is class imbalance across multiple tasks. Our 10 tasks have variable difficulty level and number of frames across episodes. We weight the 10 tasks in mid-training by the inverse frequency of their per-task frame counts for stable training. E2 post-training uses the same hyperparameters. 

\textbf{Seeds.} Our test seeds start at 1000000, validation set seeds starting at 100000000. All experiments use the same seeds. E2 tests use the same seeds. Expert demonstration from the motion planner use seeds starting at 0 across mid-training in E1 and post-training in E1 and E2. For generating self-supervised online policy rollouts for $T_{SS}$ tasks, which we save and use for rehearsal, we use seeds starting 5000. We release all test set seeds in the code.  

\textbf{Base Model Definition.} The \emph{$\pi_{0.5}$ RoboTwin Base Policy} is mid-trained on the 10 $T_{SS}$ tasks with 500 expert episodes, having 130,815 frames. It initializes all post-training experiments and generates the self-demos for \emph{Multi-Task ES+SS (Ours)}. For E2, the same checkpoint (mid-trained in E1) is rolled out in the E2 embodiment to generate the E2 self-demos.

\section{Benchmark Details}
\label{app:benchmark_details}

\begin{table}[h]
\centering
\caption{\textbf{Summary of training data and test sets across all benchmarks}. Self-demos are typically on task prompts the base policy is familiar with. \textbf{Our test sets.} $T_\text{ES}$: expert supervised tasks, $T_\text{SS}$: self-supervised tasks, $T_\text{NO}$: novel object generalization, $T_\text{NC}$: novel composition of $T_\text{ES}$ skills.}
\scriptsize
\renewcommand{\arraystretch}{1.2}
\setlength{\tabcolsep}{4pt}
\begin{tabular}{
@{}>{\centering\arraybackslash}m{1.6em}|
>{\raggedright\arraybackslash}m{1.7cm}
>{\raggedright\arraybackslash}m{4.25cm}
>{\raggedright\arraybackslash}m{4.25cm}
>{\raggedright\arraybackslash}m{1.7cm}@{}
}
\toprule
& \textbf{Split} & \textbf{Task prompts} & \textbf{Scene / objects} & \textbf{Size (Seeds)} \\
\rowcolor{gray!15}
\multicolumn{5}{@{}l@{}}{\textit{\textbf{Benchmark 1: Pick up object (real ALOHA robot, right arm only)}}} \\
\multirow[c]{2}{*}[0pt]{\rotatebox[origin=c]{90}{\textbf{TRAIN}}}
& Expert demos & Pick up the red/green cube & Red \& green cube; 8 unique cube positions & 30 \\
& Self-demos & Pick up X and place in Y; Pick up laundry place in basket & Varied objects \& scenes & 14 \\
\cmidrule(l){2-5}
\multirow[c]{4}{*}[-4pt]{\rotatebox[origin=c]{90}{\textbf{TEST}}}
& $T_\text{ES}$ & Pick up the red/green cube & 10 new positions; tested in pairs & 20 \\
& $T_\text{SS}^1$ & Pick up X and place in Y & Diverse objects X and containers Y & 20 \\
& $T_\text{SS}^2$ & Pick up laundry place in basket & 9 clothes, 2 baskets & 20 \\
& $T_\text{NO}$ & Pick up X & Novel objects or red/green cubes with novel distractors; novel scenes & 20 \\
\rowcolor{gray!15}
\multicolumn{5}{@{}l@{}}{\textit{\textbf{Benchmark 2: Gear placement (real ALOHA robot, bimanual)}}} \\
\multirow[c]{2}{*}[0pt]{\rotatebox[origin=c]{90}{\textbf{TRAIN}}}
& Expert demos & Pick up X-colored gear and place on X-colored shaft, using other arm to stabilize & 3 gear colors, $\geq$2 unique text instructions per scene & 60 \\
& Self-demos & Pick up X and place in Y; Pick up laundry place in basket & Varied objects \& scenes & 30 \\
\cmidrule(l){2-5}
\multirow[c]{2}{*}[0pt]{\rotatebox[origin=c]{90}{\textbf{TEST}}}
& $T_\text{ES}$ & Gear placement (3 training colors) & Same 3 gear colors from expert demos; new gear and board positions & 10 \\
& $T_\text{NO}$ & Gear placement (novel colors) & 3 unseen gear colors not in expert demos; new gear and board positions & 10 \\
\rowcolor{gray!15}
\multicolumn{5}{@{}l@{}}{\textit{\textbf{Benchmark 3: Block stacking (simulated RoboTwin, bimanual)}}} \\
\multirow[c]{2}{*}[0pt]{\rotatebox[origin=c]{90}{\textbf{TRAIN}}}
& Expert demos & Stack block green on red; Stack block red on green & Same block positions; motion-planned demos mirroring real setup & 50 $\times$ 2 \\
& Self-demos & 10 pretraining tasks (e.g.\ place dual shoes, open microwave, place bread basket) & Simulator rollouts of base model on stage-1 pretraining tasks & 10 $\times$ 10 \\
\cmidrule(l){2-5}
\multirow[c]{4}{*}[-4pt]{\rotatebox[origin=c]{90}{\textbf{TEST}}}
& $T_\text{ES}$ & Stack block green on red; Stack block red on green & Same task setup as expert demos & 50 $\times$ 2 \\
& $T_\text{SS}$ & All 10 pretraining tasks & Same setup as self-demo rollouts & 50 $\times$ 10 \\
& $T_\text{NO}$ & Stack cylinder on cube; Stack sphere on cube (+ w/ distractors) & Novel object shapes; random colors; same-colored distractors variant & 50 $\times$ 4 \\
& $T_\text{NC}$ & Stack bowls/blocks three with exact prompts differing from $T_\text{ES}$ & 2 bowls, 3 bowls, 3 cubes in scene & 50 $\times$ 3 \\
\bottomrule
\end{tabular}
\vspace{8pt}
\label{tab:data_splits}
\end{table}

\subsection{Success Rate Criteria}
\textbf{Pick, Place.} Since our focus is on policy's instruction following, interaction with the specified objects, and demonstrating the specified behavior, we mark success even if the policy picks or places correctly once. This is consistent across our method and the baselines. 

\textbf{Laundry.} The laundry task is challenging to distill, and we measure a partial success rate, marking success if the cloth is partially placed inside the basket. 

\textbf{Caterpillar Gear Insertion.} For a successful episode, the policy has to grasp and pick the correct colored gear, move to the specified gear shaft color and align it on top of that. We mark partial alignment as success if gear is placed on top of the correct gear. 

\textbf{Push.} Push is defined as the motion of dragging an object (push or pull) without lifting it off the table. If lifted, it is counted as pick instead. A distinction is made in actual push motion versus attempted grasp causing accidental push, which is common in pushing cups. 

\textbf{Turn/Pour.} \added{In this test set, we either ask the policy to ``turn a container over" or ``pour items from the container onto the table". The policy relies on zero-shot priors not over-written by the post-training and is typically noisier. Turning by $90^{\circ}$ or partially pouring items out is considered as having motion priors. Flipping it by $180^{\circ}$ would be too challenging. }

\subsection{RoboTwin Tasks}
\label{app:robotwin_tasks}
\begin{figure}[h]
\centering
\vspace{-0.8cm}
\includegraphics[width=\linewidth]{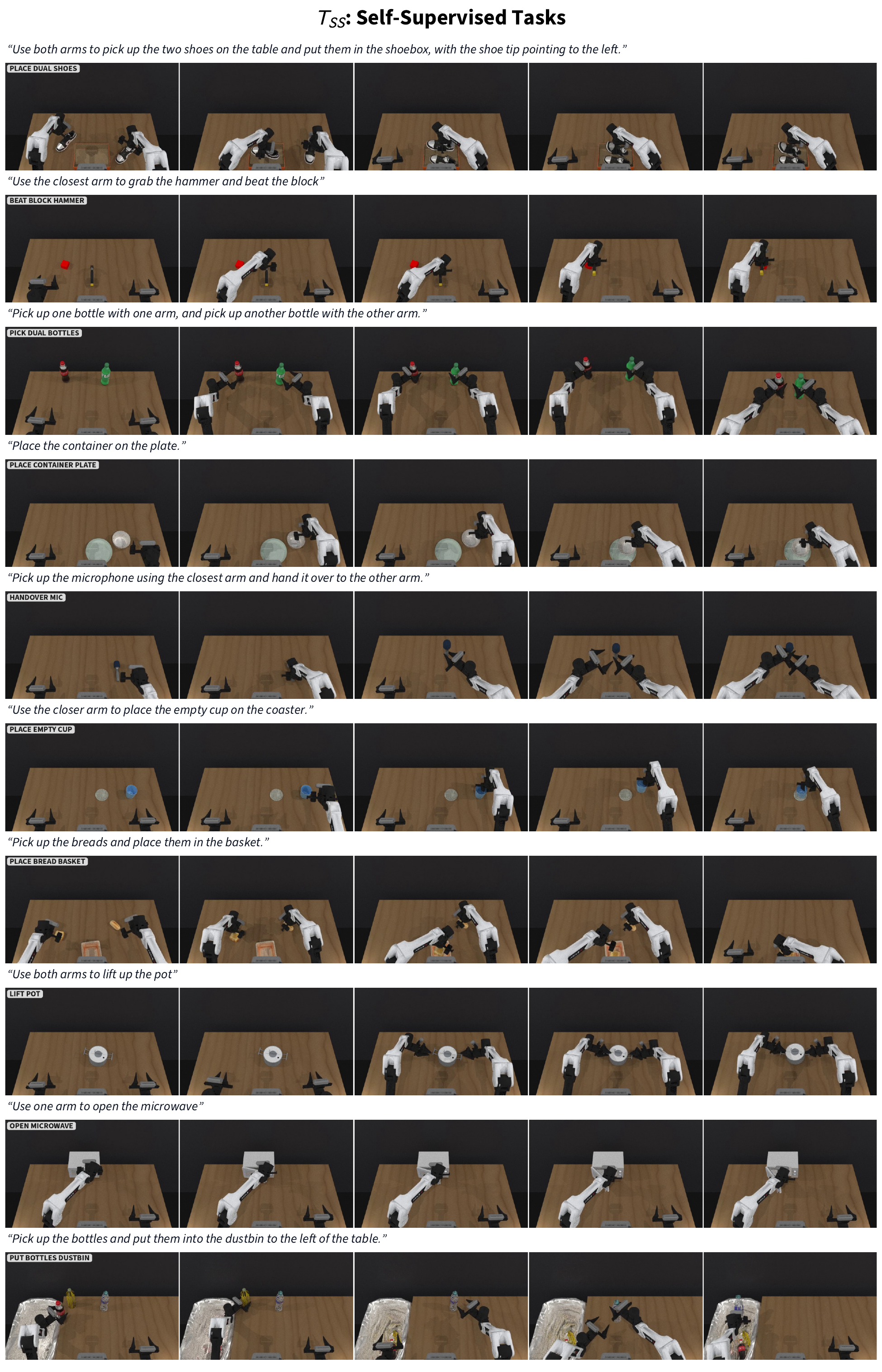}
    \caption{\textbf{$T_{SS}$ Self-supervised tasks in our RoboTwin benchmark.} These are mid-training tasks, used for rehearsal in post-training. We collect online self-demos in the simulator for post-training on these.}
    \label{fig:montage_robotwin_tss}
\vspace{-0.2cm}
\end{figure}

\begin{figure}[h]
\centering
\includegraphics[width=\linewidth]{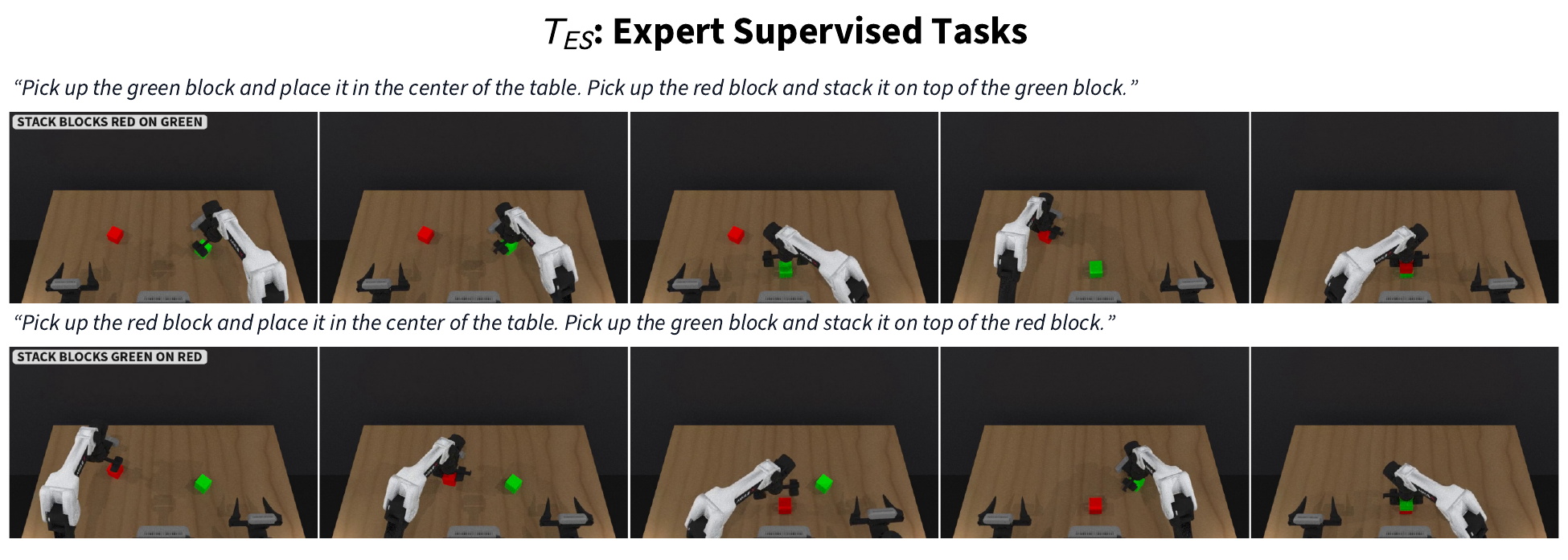}
    \caption{\textbf{$T_{ES}$ Expert-supervised tasks in our RoboTwin benchmark.} We collect motion planner expert data for these two stack block tasks in post-training.}
    \label{fig:montage_robotwin_tes}
\end{figure}

\begin{figure}[h]
\centering
\includegraphics[width=\linewidth]{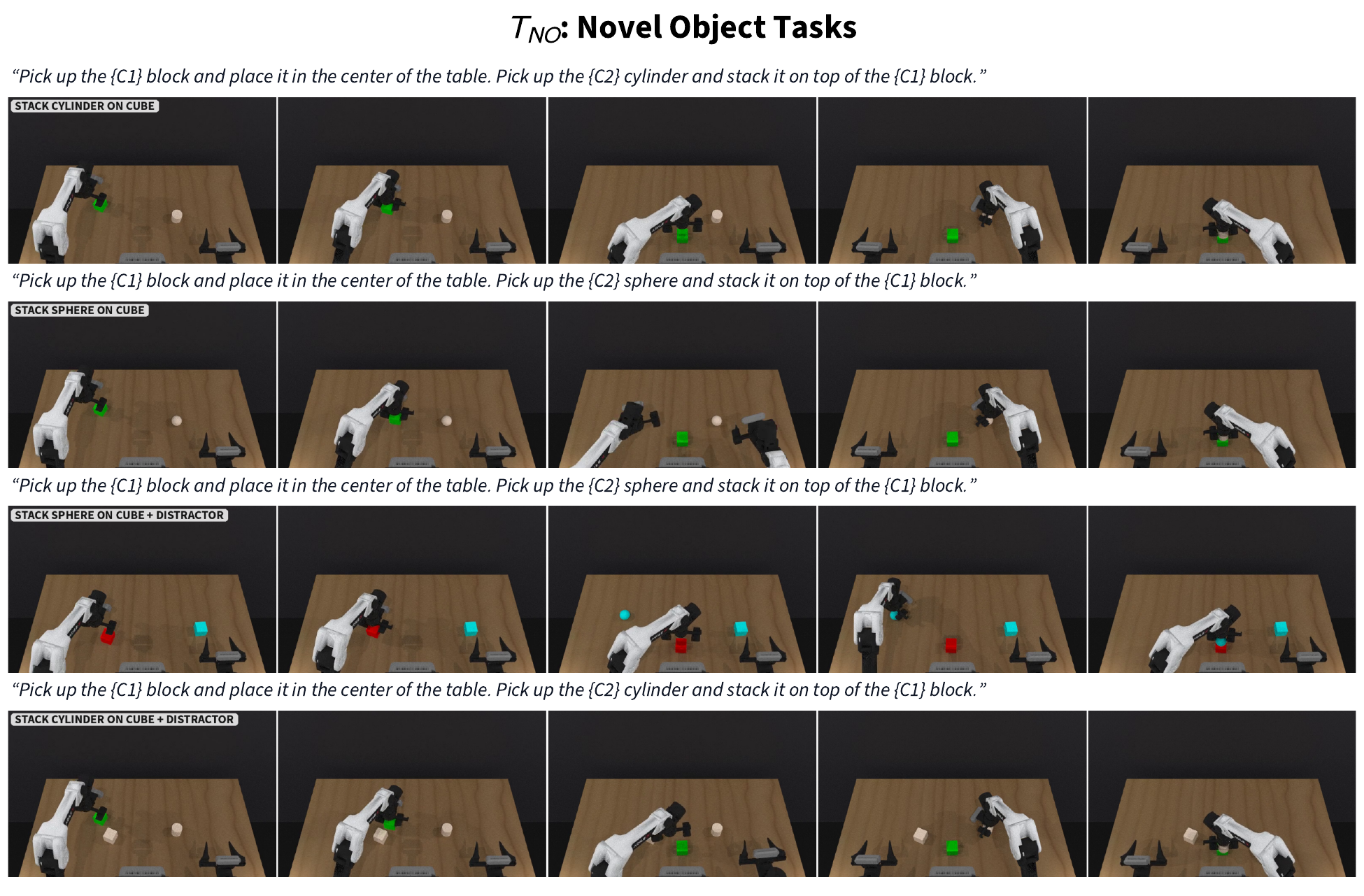}
    \caption{\textbf{$T_{NO}$ Novel-Object tasks in our RoboTwin benchmark.} These stack novel objects instead of the red and green cubes part of expert fine-tuning data.}
    \label{fig:montage_robotwin_tno}
\end{figure}

\begin{figure}[h]
\centering
\includegraphics[width=\linewidth]{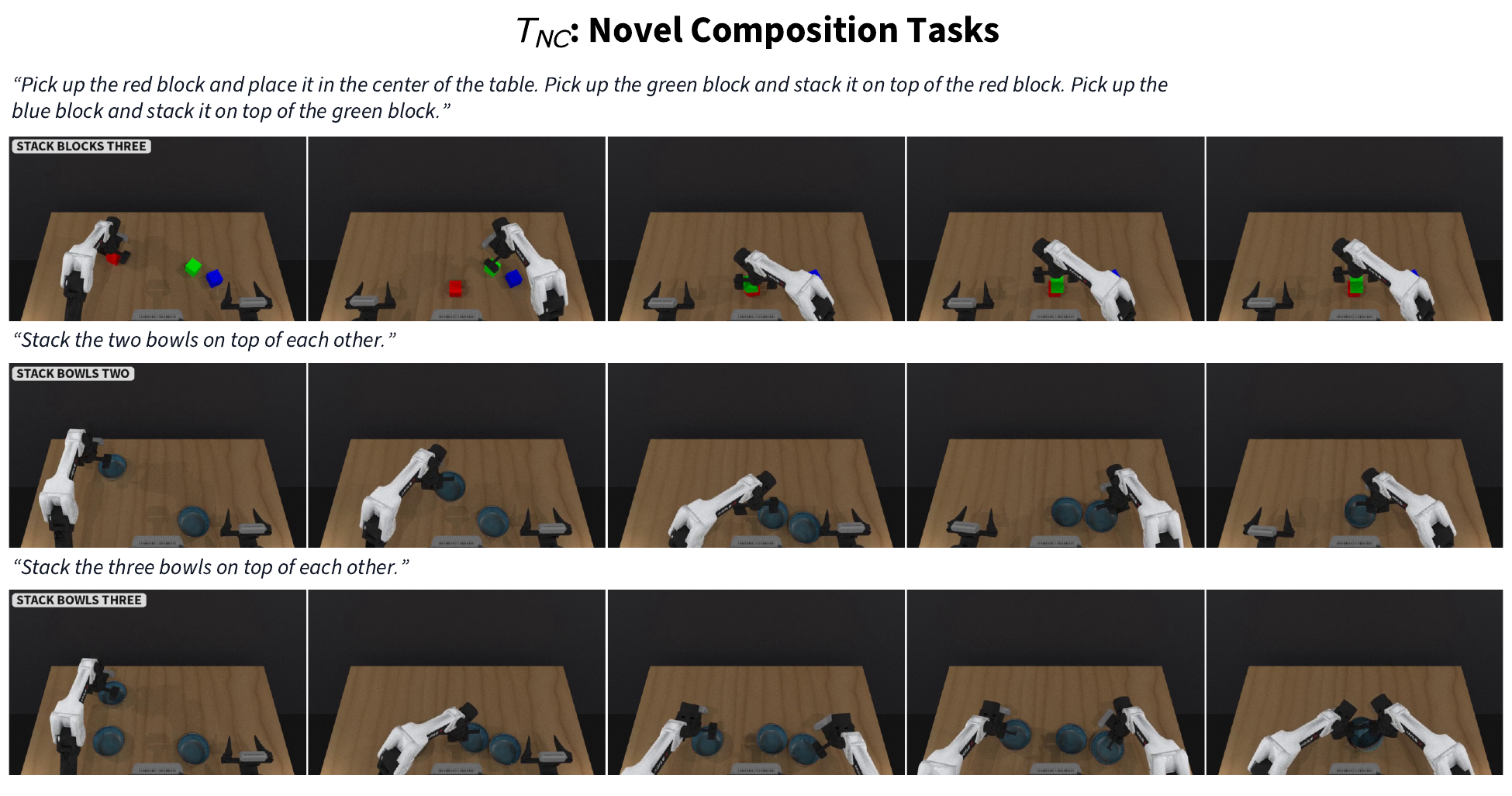}
    \caption{\textbf{$T_{NC}$ Here we compose new tasks within the skill family of stacking.} These are challenging tasks since there is no self-supervised or expert-supervised data for it. The text prompt in ``stack two bowls" is different from the text prompt in ``stack blocks" tasks in $T_{ES}$ and $T_{NO}$.}
    \label{fig:montage_robotwin_tnc}
\end{figure}

Figures~\ref{fig:montage_robotwin_tss}--\ref{fig:montage_robotwin_tnc} show representative rollouts for all four RoboTwin task splits: $T_{SS}$, $T_{ES}$, $T_{NO}$, and $T_{NC}$.

\subsection{Real ALOHA Test Sets}
\label{app:real_testsets}
\begin{figure}[h]
\centering
\includegraphics[width=\linewidth]{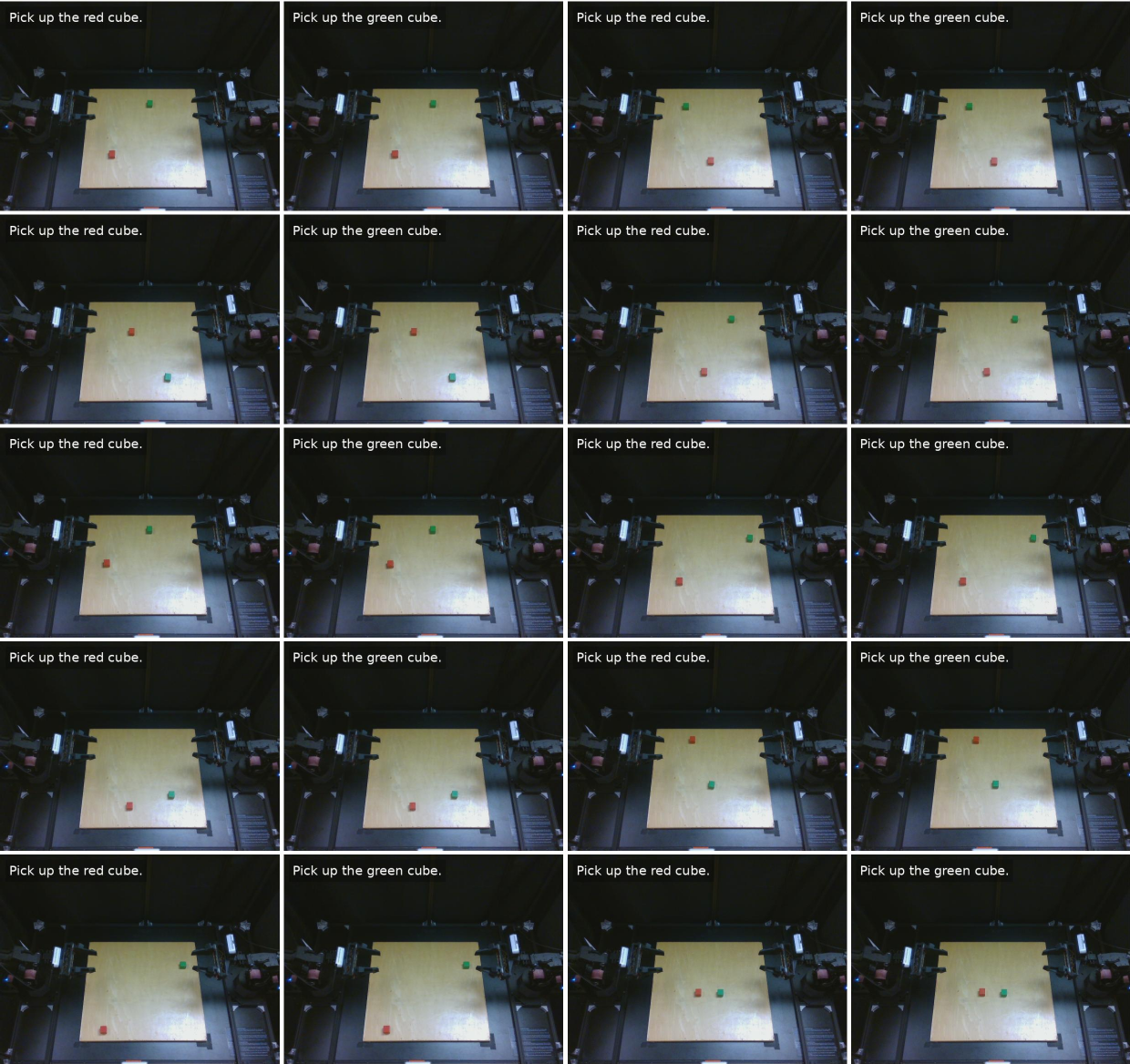}
    \caption{ALOHA test set $T_{ES}$, main paper Table \ref{tab:real_robot_results}.}
    \label{fig:test_set_ID}
\end{figure}

\begin{figure}[h]
\centering
\includegraphics[width=\linewidth]{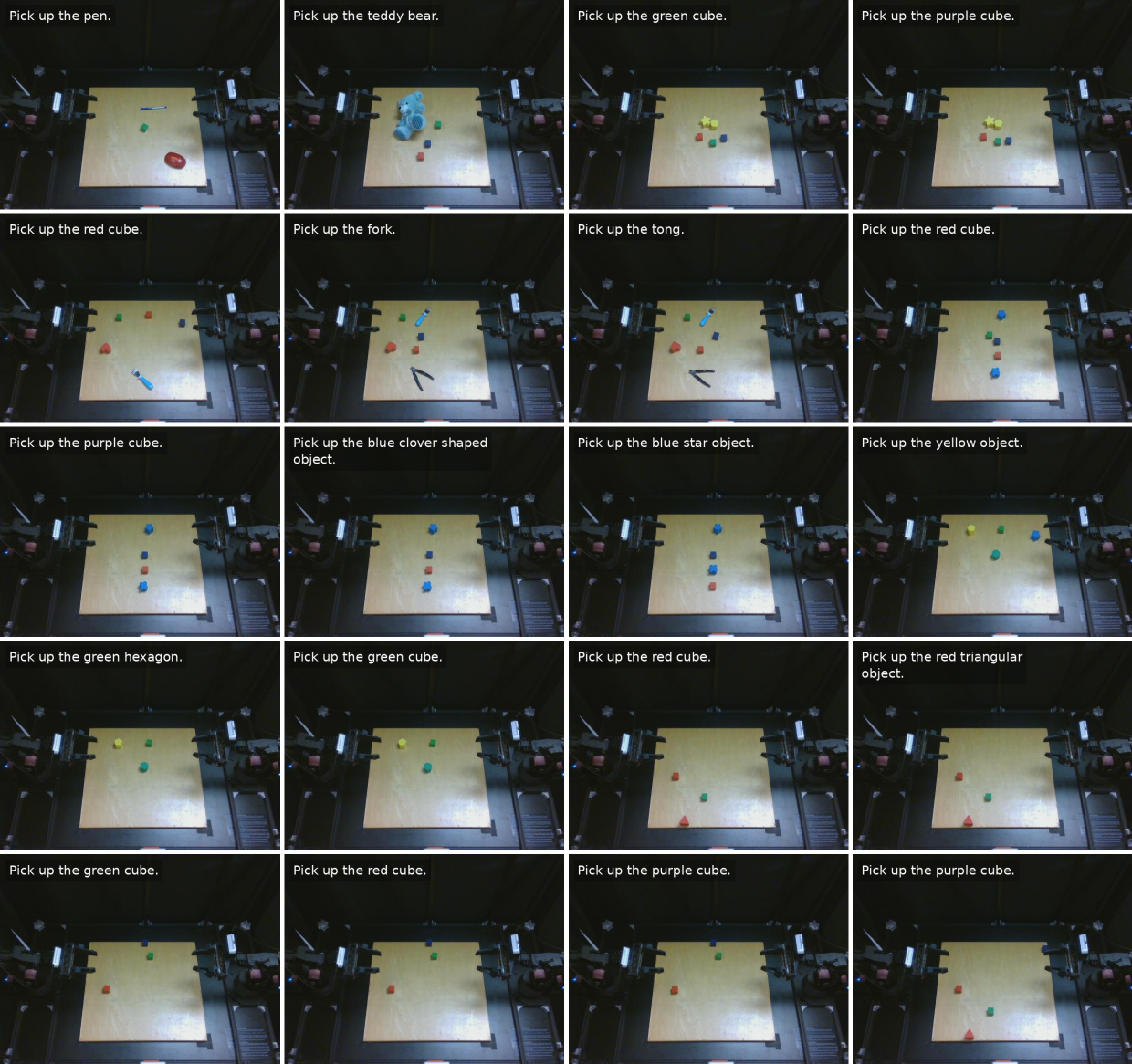}
    \caption{ALOHA test set $T_{NO}$, main paper Table \ref{tab:real_robot_results}.}
    \label{fig:test_set_OOD_object}
\end{figure}

\begin{figure}[h]
\centering
\includegraphics[width=\linewidth]{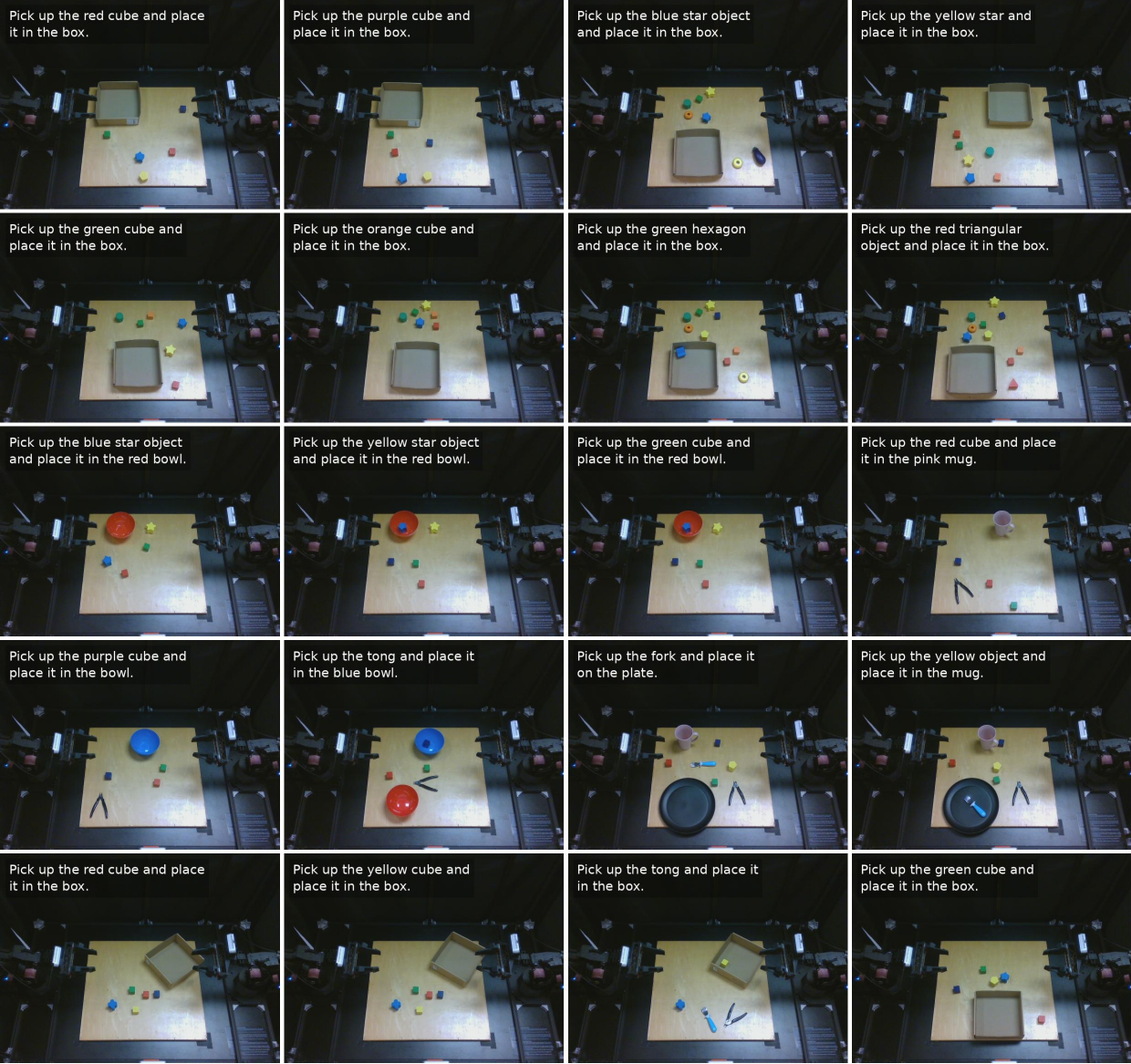}
    \caption{ALOHA test set $T_{SS}^1$, main paper Table \ref{tab:real_robot_results}.}
    \label{fig:test_set_OOD_pretrain}
\end{figure}

\begin{figure}[h]
\centering
\includegraphics[width=\linewidth]{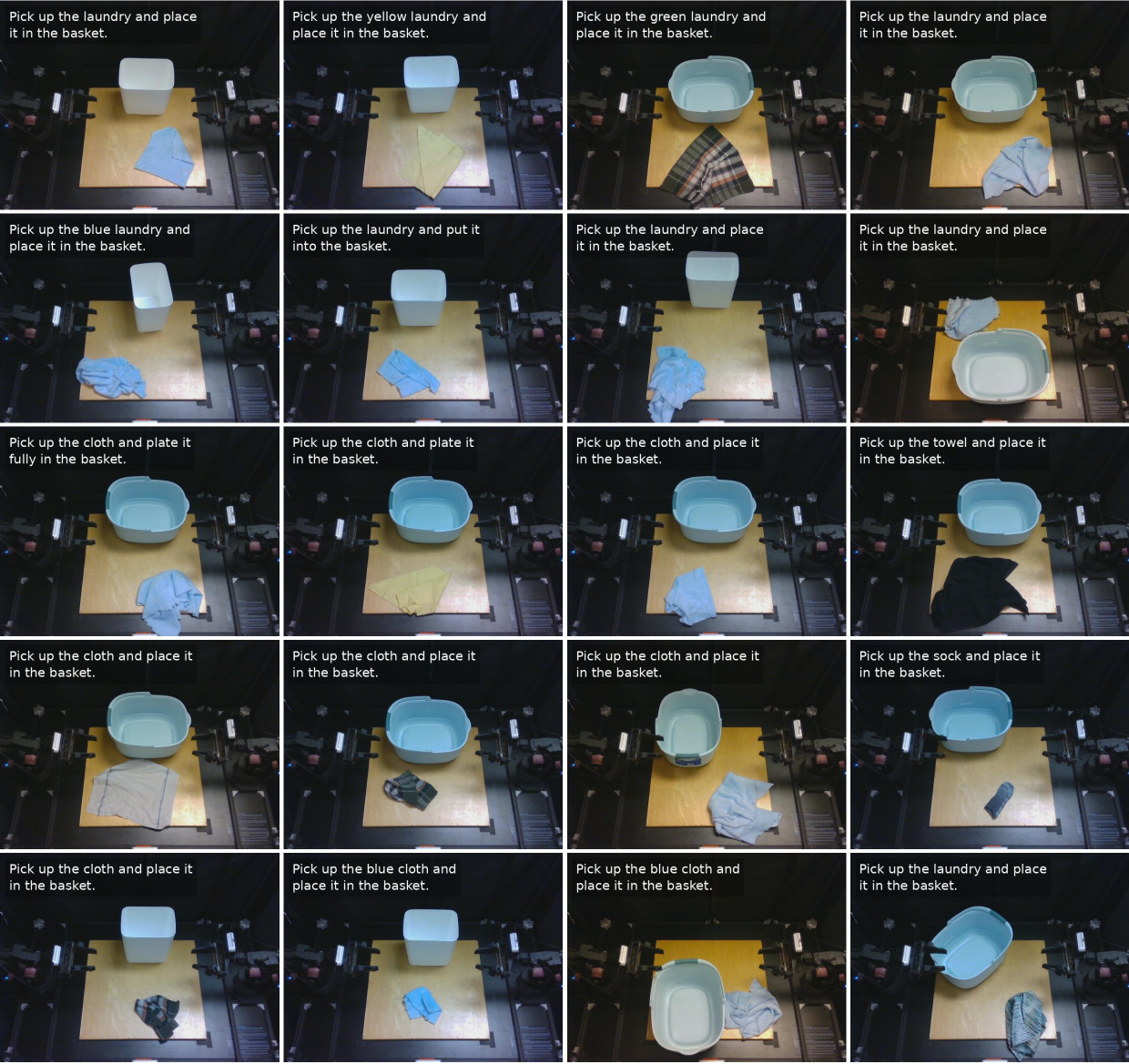}
    \caption{ALOHA test set $T_{SS}^2$, main paper Table \ref{tab:real_robot_results}.}
    \label{fig:test_set_OOD_laundry}
\end{figure}

\begin{figure}[h]
\centering
\includegraphics[width=\linewidth]{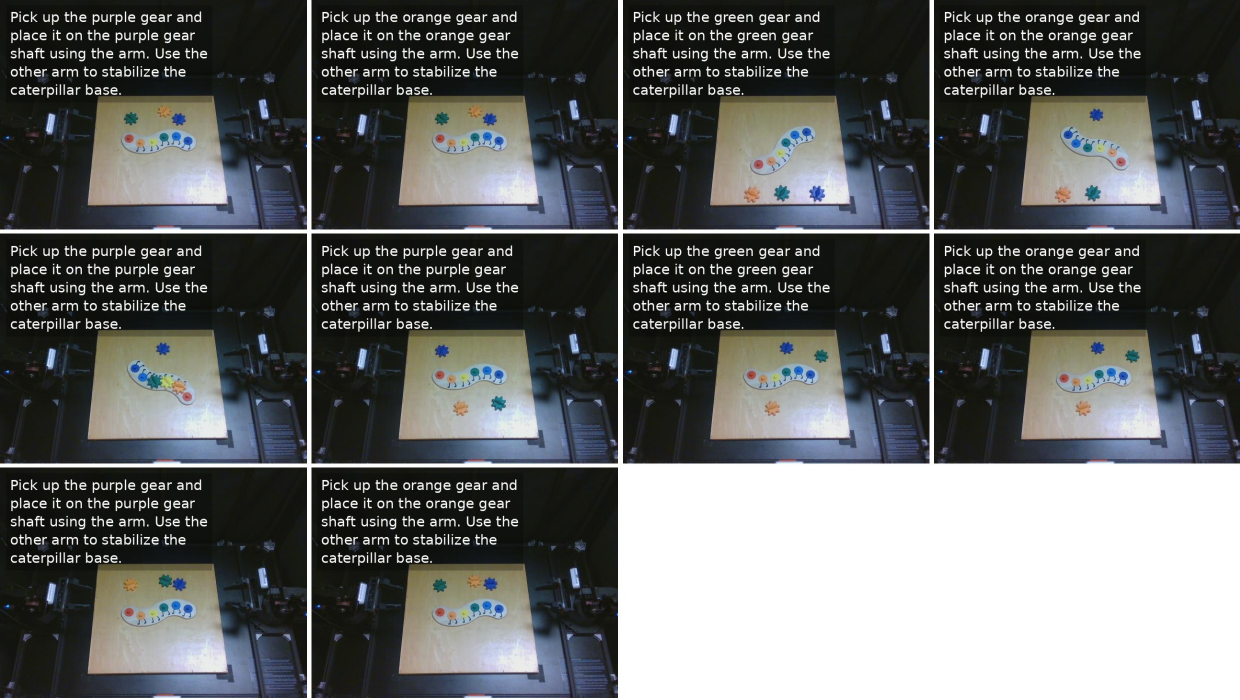}
    \caption{ALOHA test set $T_{ES}$, main paper Table \ref{tab:caterpillar}.}
    \label{fig:test_set_ID_caterpillar}
\end{figure}

\begin{figure}[h]
\centering
\includegraphics[width=\linewidth]{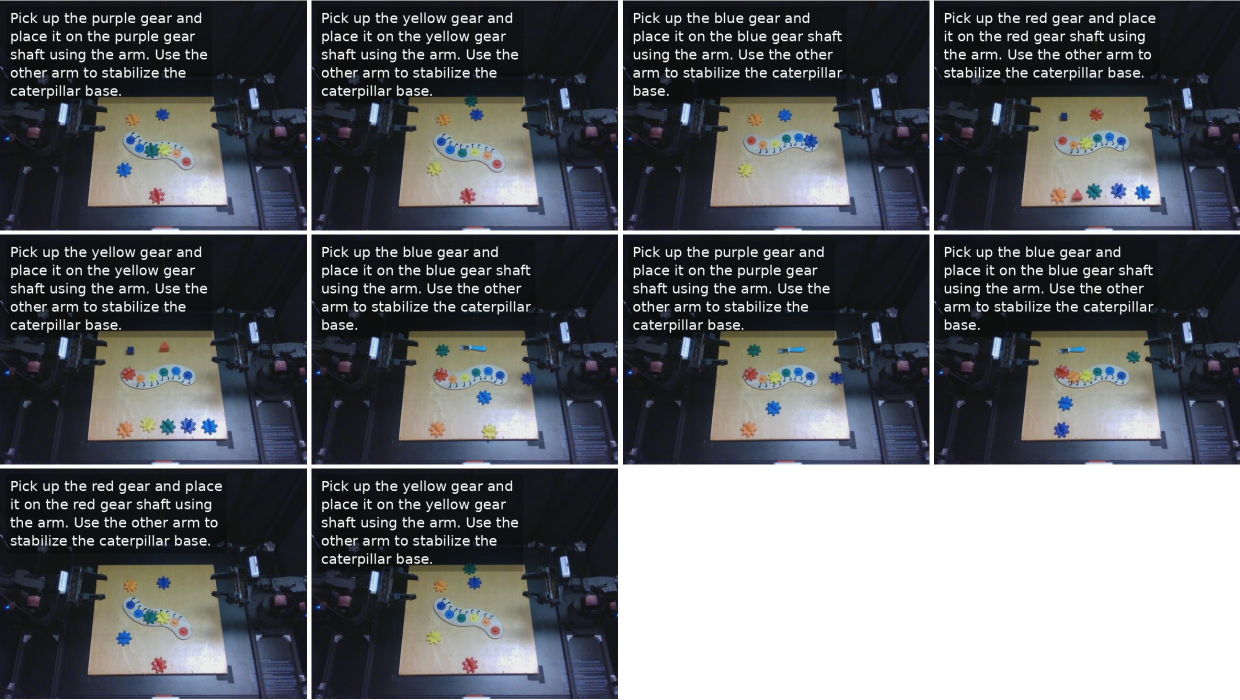}
    \caption{ALOHA test set $T_{NO}$, main paper Table \ref{tab:caterpillar}.}
    \label{fig:test_set_OOD_caterpillar}
\end{figure}

For reproducibility, Figures~\ref{fig:test_set_ID}--\ref{fig:test_set_OOD_laundry} visualize the four real ALOHA test sets ($T_{ES}$, $T_{NO}$, $T_{SS}^1$, $T_{SS}^2$) used to evaluate pick-up-only and pick-and-place tasks in Table~\ref{tab:real_robot_results}, and Figures~\ref{fig:test_set_ID_caterpillar}--\ref{fig:test_set_OOD_caterpillar} show the $T_{ES}$, $T_{NO}$ test scenes for the caterpillar gear placement benchmark in Table~\ref{tab:caterpillar}. 

\begin{figure}[h]
\centering
\includegraphics[width=\linewidth]{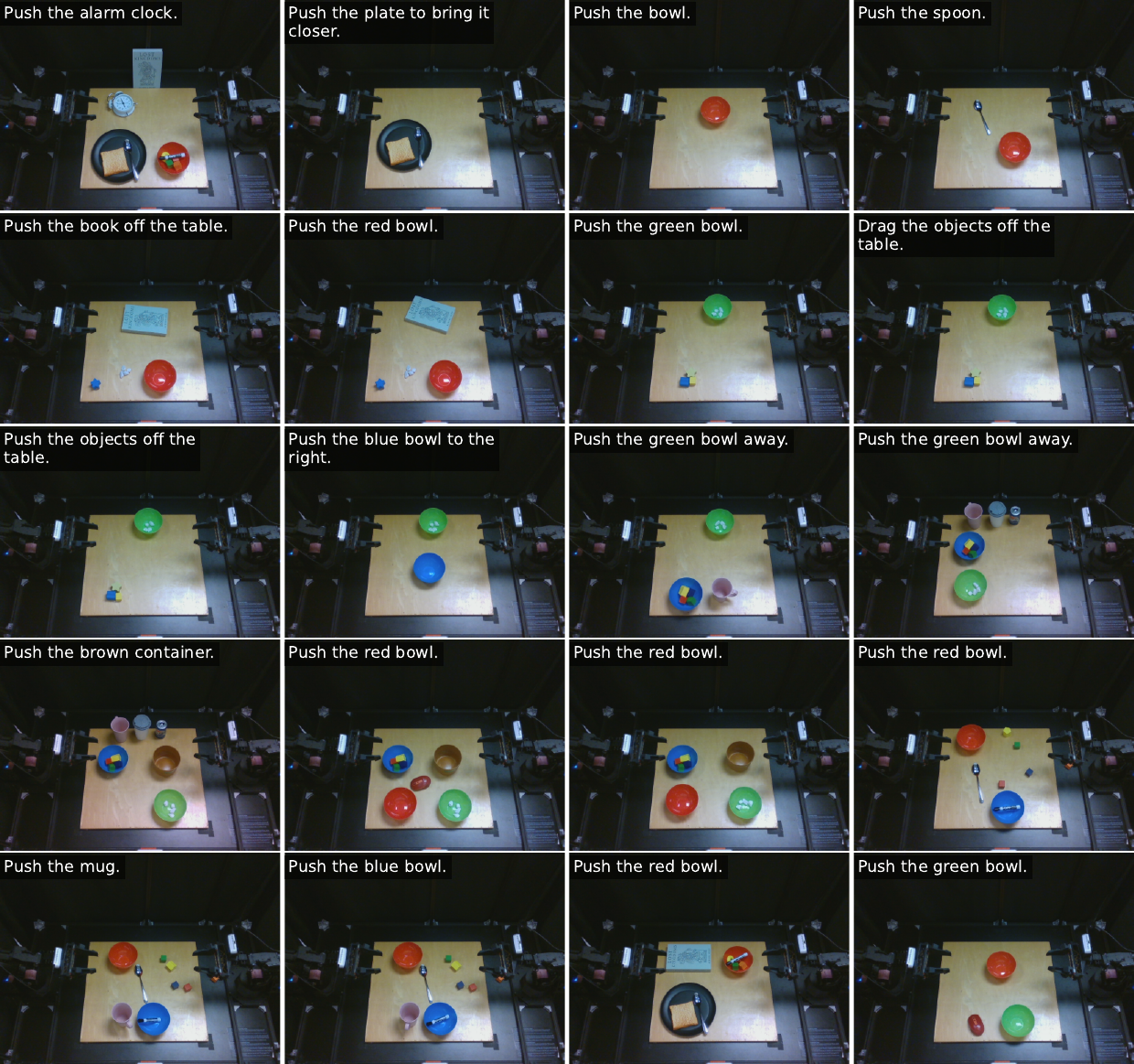}
    \caption{ALOHA test set, held-out skill family ``Push".}
    \label{fig:test_Set_push}
\end{figure}

\begin{figure}[h]
\centering
\includegraphics[width=\linewidth]{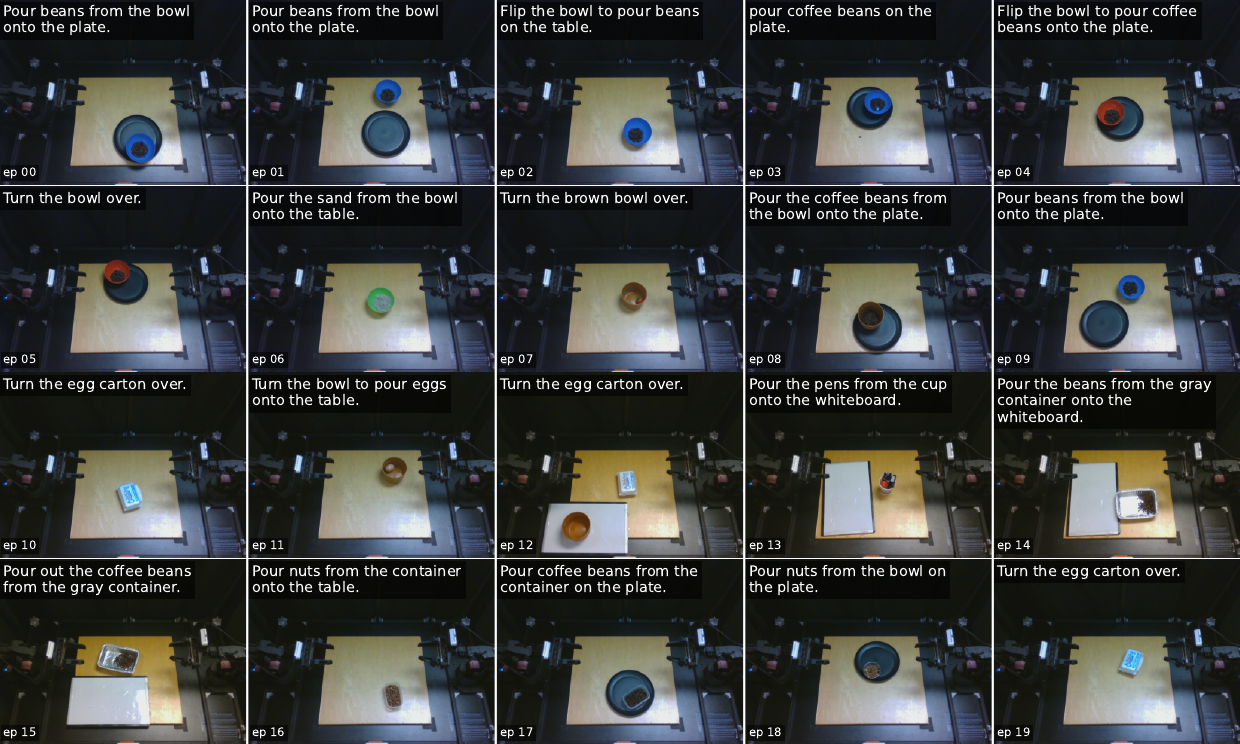}
    \caption{ALOHA test set, held-out skill family ``Turn/Pour".}
    \label{fig:test_set_pour}
\end{figure}

\section{ALOHA Robot Rollouts and Analysis}
\label{app:real_rollouts}
\begin{figure}[p]
\centering
\includegraphics[width=\linewidth]{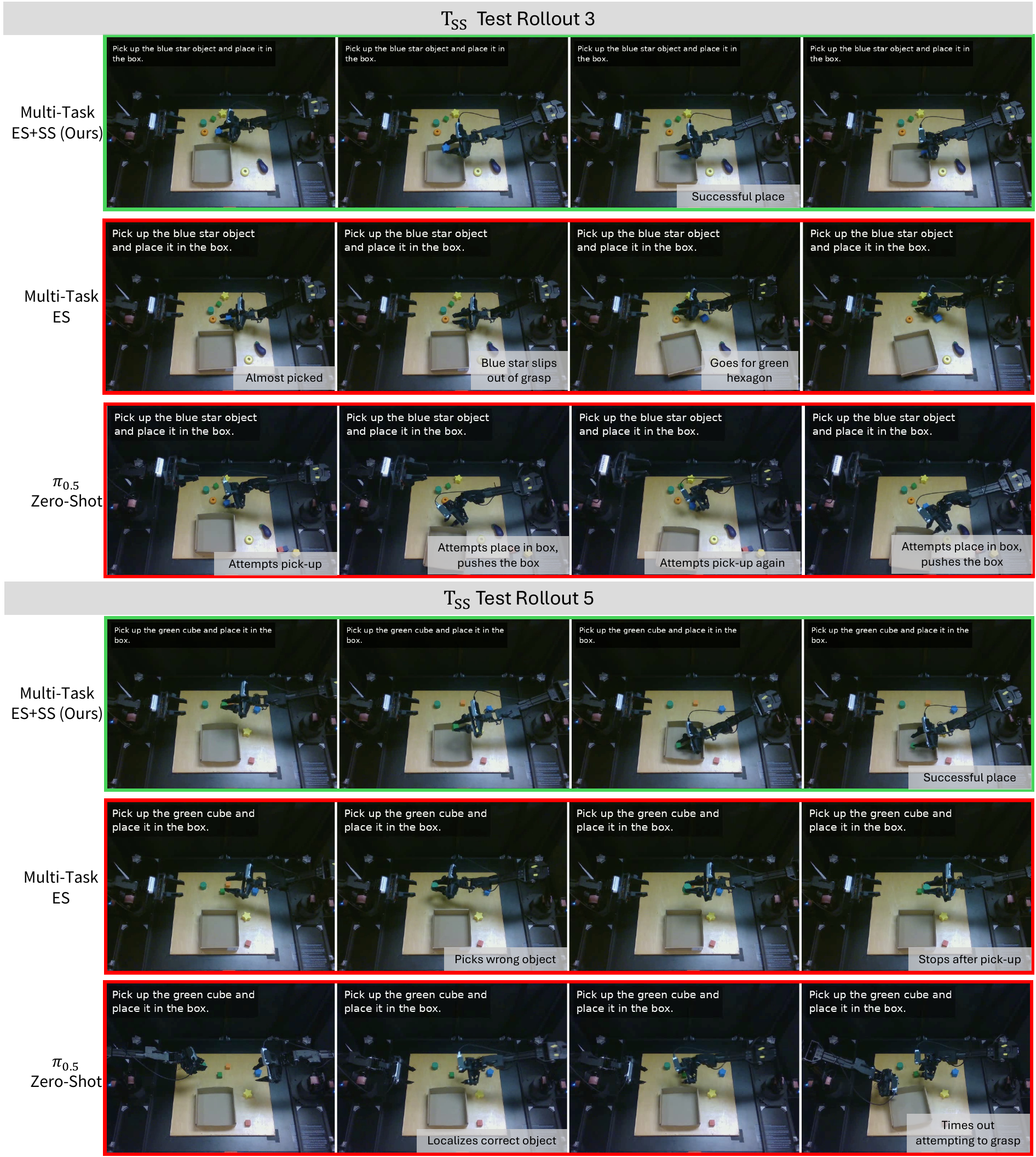}
    \caption{This figure demonstrates rollouts comparing our method with the \emph{Multi-Task ES} and \emph{zero-shot $\pihalf$} baselines. Even when the base policy shows correct instruction following and semantic understanding, it can't grasp objects due to a potential calibration gap on our embodiment. \emph{Multi-Task ES} baseline picks up the wrong object and stops, whereas our \emph{Multi-Task ES+SS} has stronger recovery behavior to pick up the correct object and successfully places objects.}
    \label{fig:montage_ood_pretrain_pick_R_1}
\end{figure}

\begin{figure}[p]
\centering
\includegraphics[width=\linewidth]{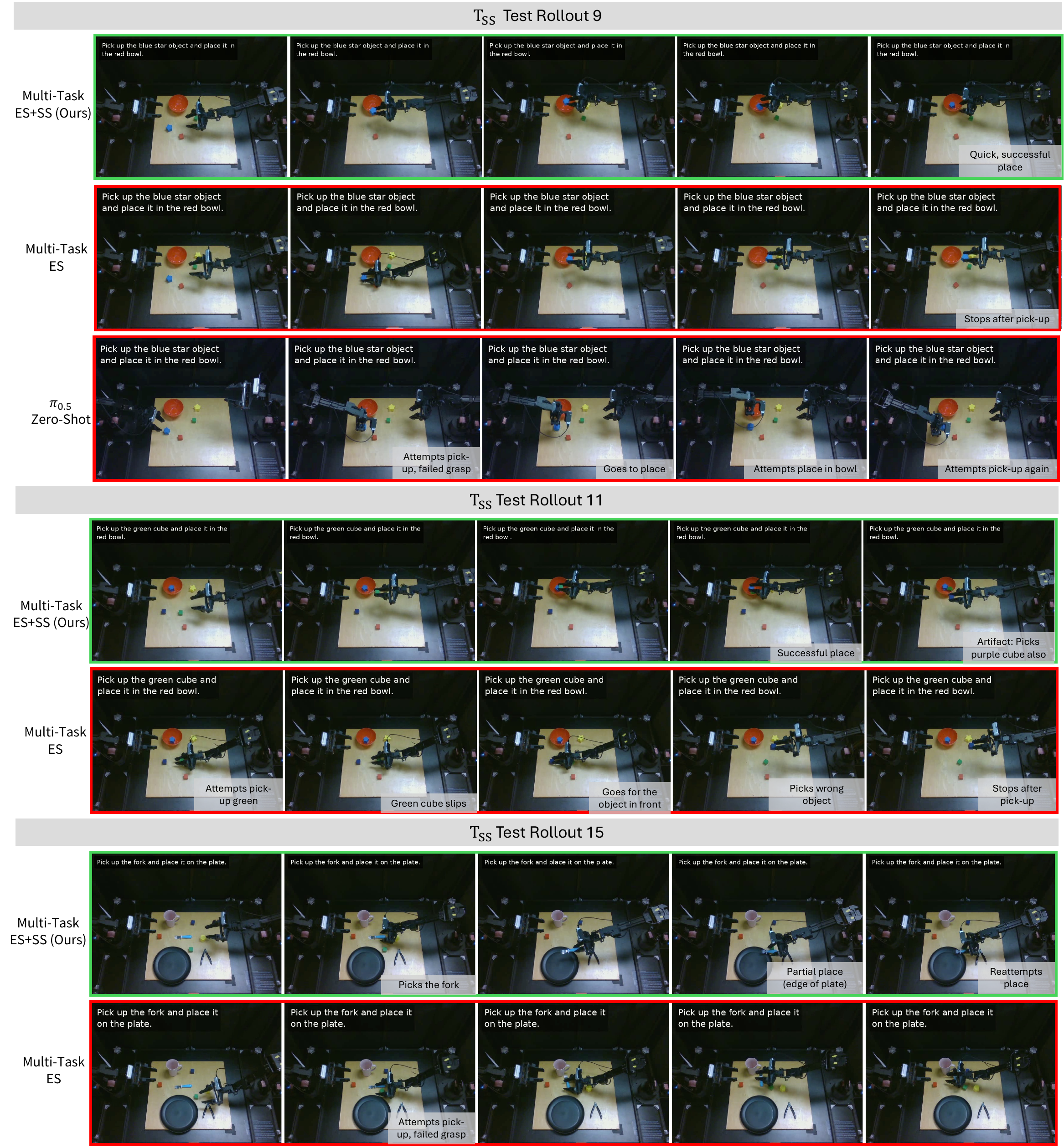}
    \caption{Our method is able to better localize and grasp the fork (rollout 15), does not stop after pick up (rollout 9). In rollout 11, it successfully places the green cube but also goes for the purple afterwards which is an artifact we observed.}
    \label{fig:montage_ood_pretrain_pick_R_2}
\end{figure}

\begin{figure}[p]
\centering
\vspace{-1.5cm}
\includegraphics[width=\linewidth]{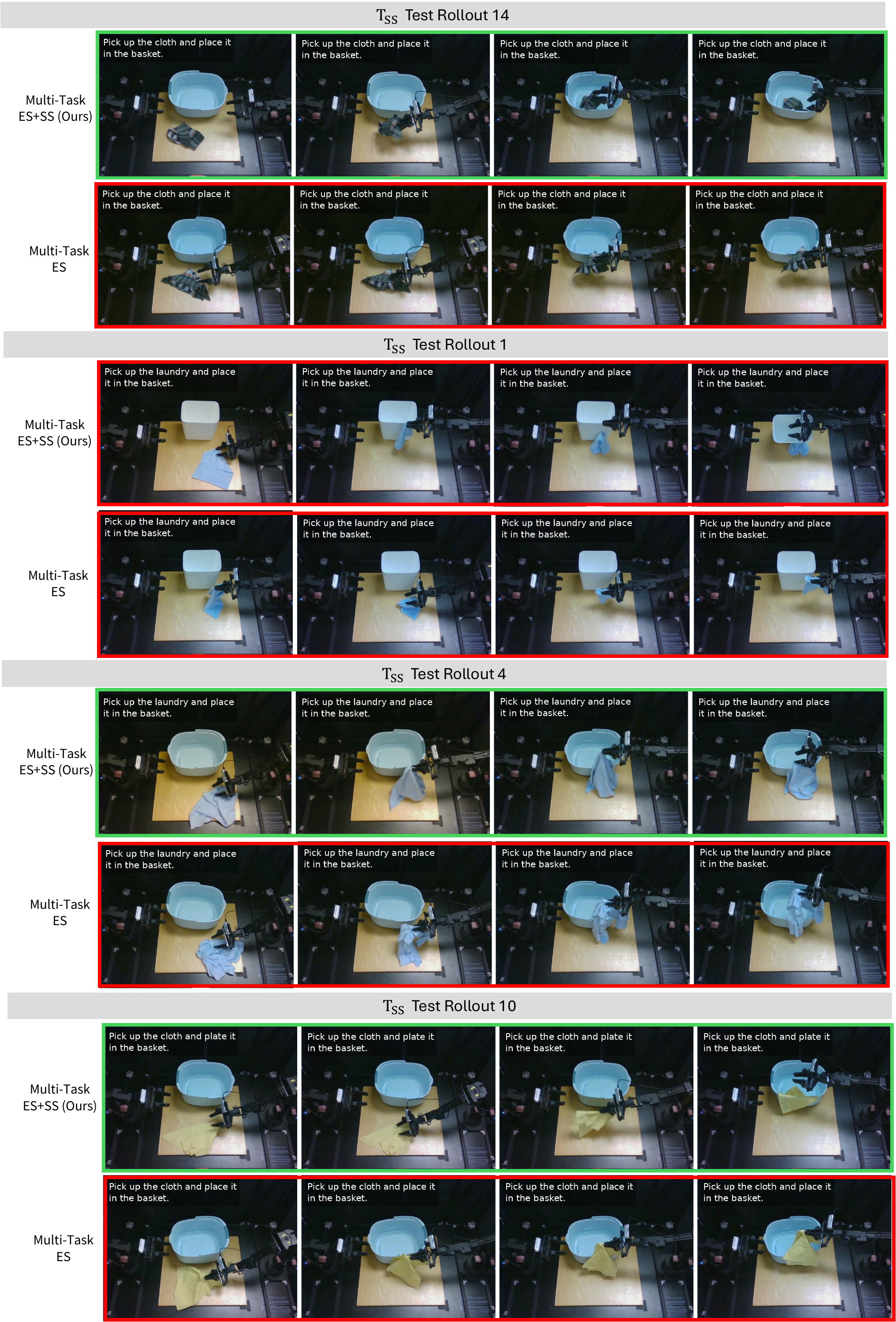}
    \caption{In these rollouts, we show the last frame of the test episode in the last column. This is a particularly challenging task to even distill because the base policy rollouts are very noisy. The common failure mode for the \emph{Multi-Task ES} baseline with expert-only `Pick up X’ data is that even after successful grasps, it overfits to the `Pick up’ task and stops (episodes 1 and 14). In some cases, it moves towards the basket (episodes 4 and 10) but even if its directly above the basket, it does not release the cloth. This is consistent with the `Pick up X and place in container Y’ test set and the baseline cannot tell the difference between pick up and pick up and place. Our method on the other hand, goes above the edge of the basket and releases the cloth. Due to the difficulty of the task and our focus on instruction following, we consider partial success (such as any part of the cloth placed inside the basket when the episode ends) as a success.}
    \label{fig:montage_ood_pretrain_laundry}
\vspace{-0.5cm}
\end{figure}

\begin{figure}[p]
\centering
\includegraphics[width=\linewidth]{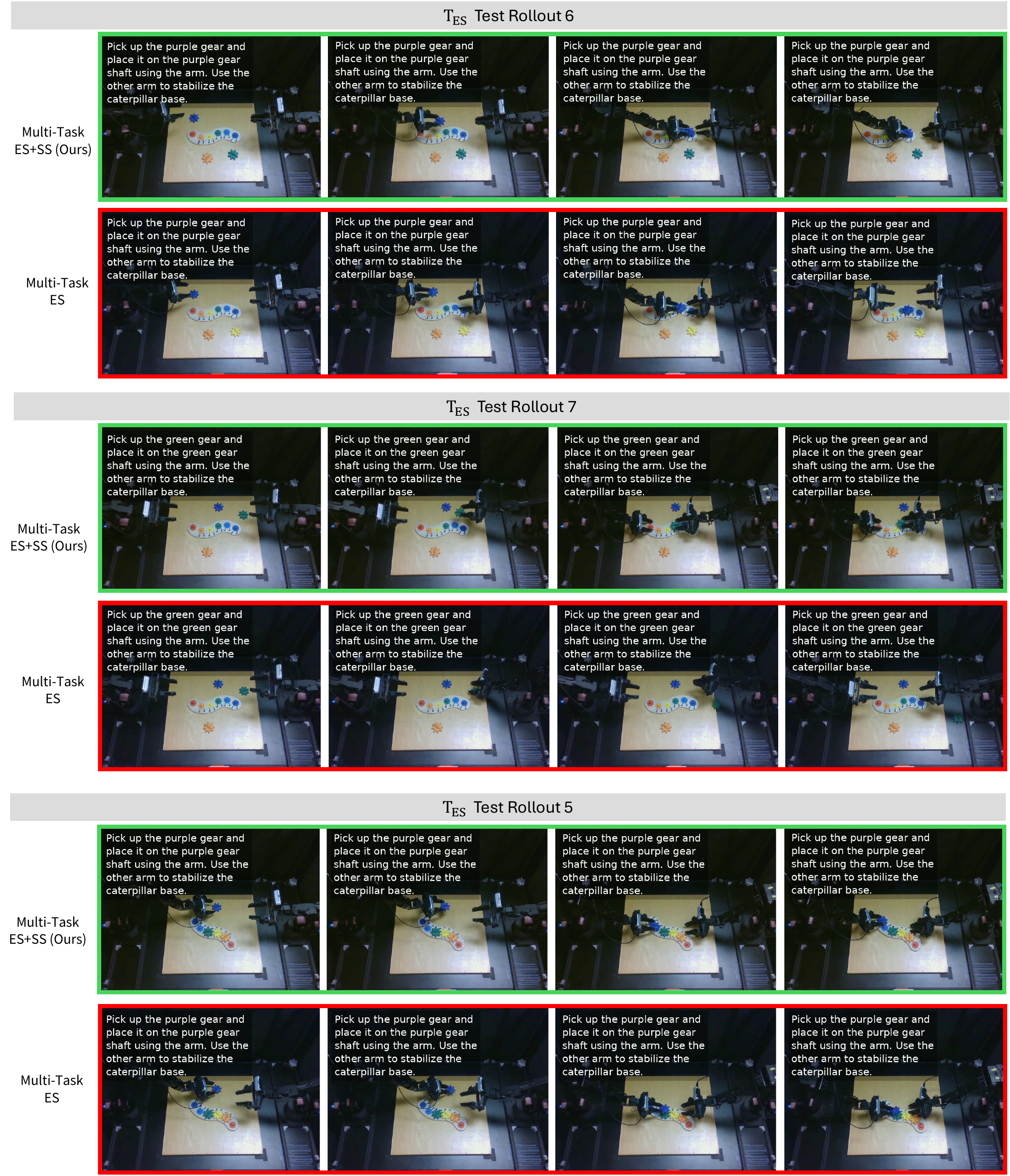}
    \caption{\textbf{Caterpillar Benchmark, $T_{ES}$ test set rollouts.} [Rollout 6] The purple and blue colors are hard to tell and our method is able to do that (notice the purple gear is placed on purple shaft in ours and on the blue shaft in the baseline). [Rollout 7] Several of the failures of the baseline is due to unstable grasps, resulting in the gears slipping out and being thrown far away with no recovery behavior. [Rollout 5] Our method placed the purple gear on the purple shaft as compared to the baseline which placed it on top of the green shaft.}
    \label{fig:montage_caterpillar}
\end{figure}

\begin{figure}[b]
\centering
\vspace{-1.5cm}
\includegraphics[width=\linewidth]{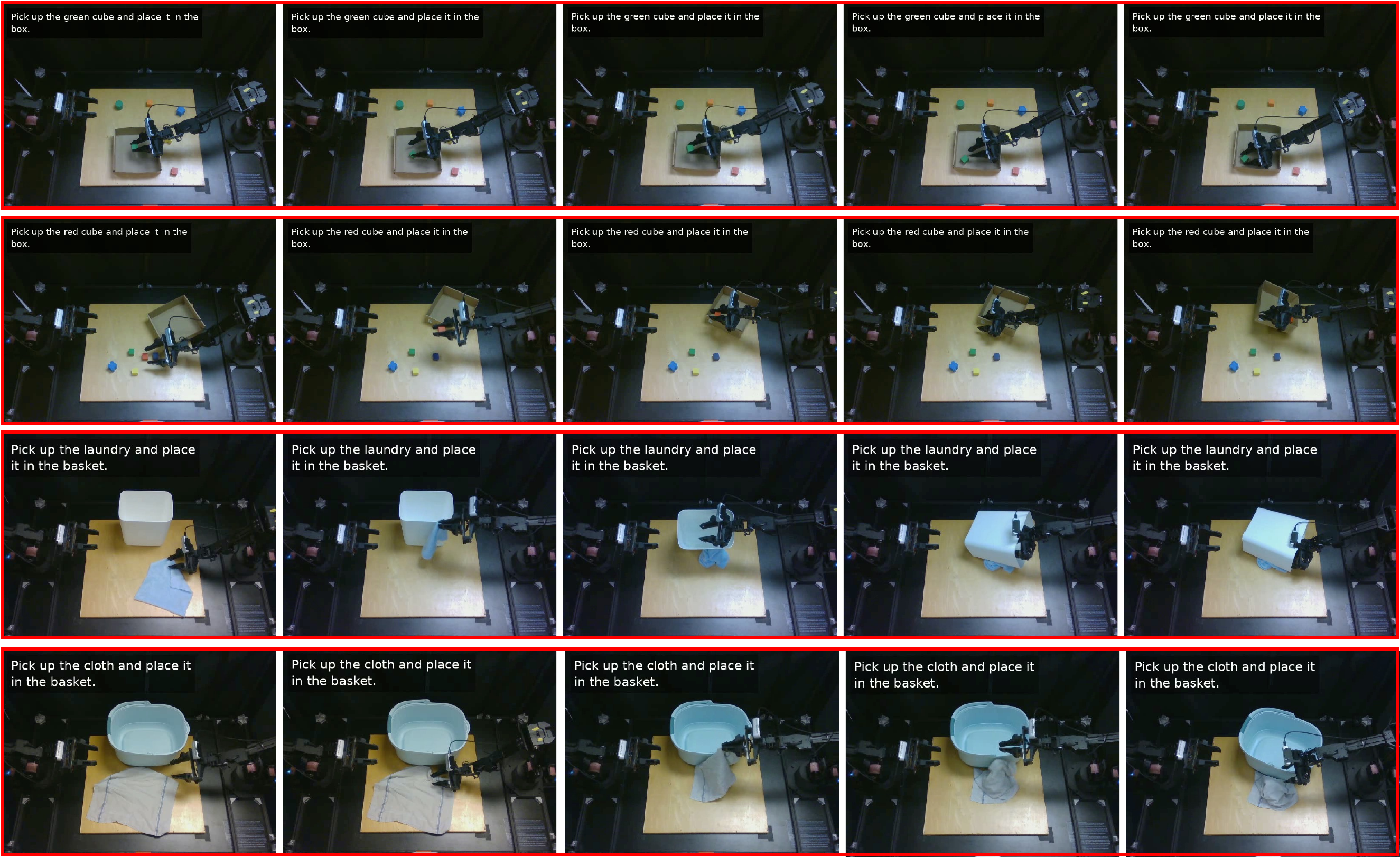}
    \caption{Here, we discuss four \textbf{failure cases} of our method. \textcolor{blue}{[Row 1]} This is an example when the robot repeatedly picks and places the object inside the box, after successfully picking and placing it from the table. This is the failure mode we discussed in Section \ref{sec:ablation-analysis} of main paper, but it can also be seen as correct behavior since we don’t specify that the policy not pick it up if its inside the box. \textcolor{blue}{[Row 2]} After a successful pick and place of the red cube, the robot pushes and lifts the edge of the box. This is a kind of noisy behavior we observed in the self-demos, which is being distilled into our fine-tuned policy. \textcolor{blue}{[Row 3]} The laundry self-demos can be unsafe with a majority of them grabbing the edge of the container and sometimes exhibiting behavior unsafe for the wrist camera as it topples the basket and tries to go inside it. \textcolor{blue}{[Row 4]} The most common failure mode we observe for laundry is that it loosens its grasp or releases the cloth at the edge of the basket resulting in it falling outside. Having seen no expert data for pick and place, these tasks are limited by the base policy’s distilled failure behavior and can be iteratively improved.}
    \label{fig:montage_failure_mode_rollouts}
\vspace{-0.5cm}
\end{figure}

Figures~\ref{fig:montage_ood_pretrain_pick_R_1}, \ref{fig:montage_ood_pretrain_pick_R_2}, \ref{fig:montage_ood_pretrain_laundry} present real-robot rollouts on the ALOHA platform across self-supervised pick-and-place and laundry tasks from Table \ref{tab:real_robot_results}. Detailed per-rollout analysis is provided in each figure caption. These are all tasks from a single model, fine-tuned with only 2.2 minutes of human teleoperation data. Figure \ref{fig:montage_caterpillar} shows caterpillar task rollouts from Table \ref{tab:caterpillar}. All qualitative rollout images are captured from the top camera during test set evaluations (left and right eye-in-hand cameras skipped for brevity).

\end{document}